\documentclass[dvipsnames, format=sigconf, anonymous=false, review=false, nonacm=true]{acmart} 

\AtBeginDocument{%
  \providecommand\BibTeX{{%
    \normalfont B\kern-0.5em{\scshape i\kern-0.25em b}\kern-0.8em\TeX}}}

\copyrightyear{2024}
\acmYear{2024}
\setcopyright{rightsretained}
\acmConference[GECCO '24]{Genetic and Evolutionary Computation Conference}{July 14--18, 2024}{Melbourne, VIC, Australia}
\acmBooktitle{Genetic and Evolutionary Computation Conference (GECCO '24), July 14--18, 2024, Melbourne, VIC, Australia}
\acmDOI{10.1145/3638529.3654089}
\acmISBN{979-8-4007-0494-9/24/07}

\newcommand{\longname}[0]{MAP-Elites-Multi-ES}
\newcommand{\name}[0]{MEMES}
\newcommand{\optim}[0]{N_{optim}}
\newcommand{\archive}[0]{\mathcal{A}}
\newcommand{\noveltyarchive}[0]{\mathcal{N}}
\newcommand{\nsamples}[0]{M}
\newcommand{\stagnatename}[0]{stale counter}
\newcommand{\stagnate}[0]{S}
\newcommand{\stagnatebudgetname}[0]{stale budget}
\newcommand{\stagnatebudget}[0]{S_{max}}

\usepackage[ruled, lined, linesnumbered, commentsnumbered, noend]{algorithm2e}
\usepackage{xcolor}
\usepackage{soul}  

\usepackage{enumitem}
\usepackage{microtype}
\begin{document}

\title{
Enhancing MAP-Elites with Multiple Parallel Evolution Strategies
}

\author{Manon Flageat}
\authornote{Both authors contributed equally to this research.}
\email{manon.flageat18@imperial.ac.uk}
\affiliation{%
  \institution{Imperial College London}
  \city{London}
  \country{U.K.}
}

\author{Bryan Lim}
\email{bwl116@ic.ac.uk}
\authornotemark[1]
\affiliation{%
  \institution{Imperial College London}
  \city{London}
  \country{U.K.}
}

\author{Antoine Cully}
\email{a.cully@imperial.ac.uk}
\affiliation{%
  \institution{Imperial College London}
  \city{London}
  \country{U.K.}
}

\renewcommand{\shortauthors}{Manon Flageat*, Bryan Lim* and Antoine Cully}

\begin{abstract}

With the development of fast and massively parallel evaluations in many domains, Quality-Diversity (QD) algorithms, that already proved promising in a large range of applications, have seen their potential multiplied. 
However, we have yet to understand how to best use a large number of evaluations as using them for random variations alone is not always effective. 
High-dimensional search spaces are a typical situation where random variations struggle to effectively search. 
Another situation is uncertain settings where solutions can appear better than they truly are and naively evaluating more solutions might mislead QD algorithms.
In this work, we propose \longname{} (\name{}), a novel QD algorithm based on Evolution Strategies (ES) designed to exploit fast parallel evaluations more effectively.
\name{} maintains multiple (up to $\sim 100$) simultaneous ES processes, each with its own independent objective and reset mechanism designed for QD optimisation, all on just a single GPU. 
We show that \name{} outperforms both gradient-based and mutation-based QD algorithms on black-box optimisation and QD-Reinforcement-Learning tasks, demonstrating its benefit across domains. 
Additionally, our approach outperforms sampling-based QD methods in uncertain domains when given the same evaluation budget.
Overall, \name{} generates reproducible solutions that are high-performing and diverse through large-scale ES optimisation on easily accessible hardware.

\end{abstract}



\begin{teaserfigure}
    \centering
  \includegraphics[width=0.97\textwidth]{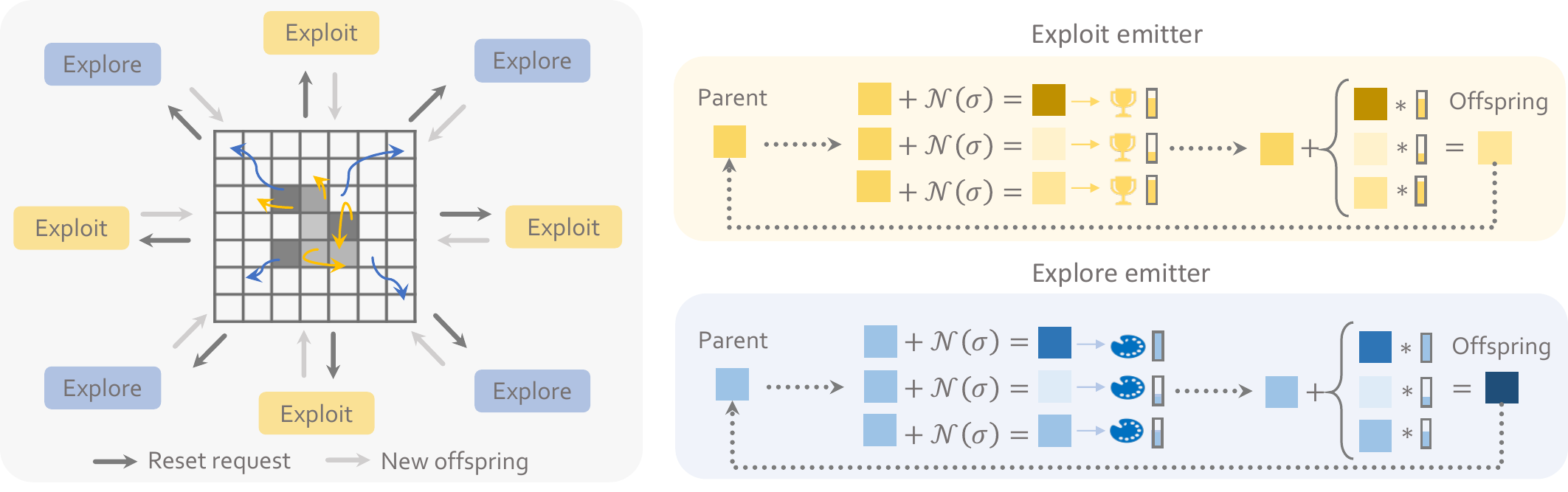}
  \caption{
    \name{} maintains multiple independent ES emitters to update a MAP-Elites archive. 
    Part of the emitters optimise for the task fitness (exploit emitter) and the rest for the novelty score (explore emitter).
    Each emitter can be independently reset based on its ability to improve the archive; when reset, an emitter restarts from a uniformly sampled solution of the archive.
  }
  \label{fig:featurefig}
\end{teaserfigure}

\maketitle

\section{Introduction}

Advances in Quality-Diversity (QD) have highlighted the benefit of diversity in optimisation. 
Maintaining a collection of diverse and high-performing solutions has proven beneficial across a wide range of applications, from robotics~\cite{cully2015robots} to design~\cite{gaier2018data} and content generation~\cite{gravina2019procedural}. 
QD algorithms help improve exploration during optimisation~\cite{ecoffet2021first, chalumeau2022assessing}, find better-performing solutions by maintaining stepping stones~\cite{mouret2015illuminating, gaier2019quality}, and facilitate rapid adaptation to unknown downstream tasks~\cite{cully2015robots}.

Previously, a large number of evaluations in QD algorithms implied costly compute resources as evaluations had to be parallelised across many CPU devices in most applications, causing a major bottleneck.
However, advances in optimisation and machine learning have been facilitated by advances in compute and hardware acceleration. 
Evolutionary Algorithms, population-based approaches and QD algorithms are not to be outdone, and are shown to be ideally positioned to take advantage of these advances. 
Recent libraries such as EvoJax~\cite{evojax2022}, QDax~\cite{chalumeau2023qdax} and EvoSax~\cite{evosax2022github} allow a speed-up in run-times by several orders of magnitude thanks to the use of modern hardware accelerators such as GPUs or TPUs. 
The vectorisation of many computational operations on these devices leads to high parallelisation of evaluations.
The development of fast highly-parallel simulators and domains~\cite{brax2021github, makoviychuk2021isaac, evojax2022, bonnet2023jumanji, jiang2023minimax, flair2023jaxmarl} also enables these advances to be applied in a wide range of both simple and complex applications.
Lim et al.~\cite{lim2022accelerated} demonstrate that QD algorithms, and in particular MAP-Elites, scale extremely well to massive parallelisation of evaluations.
Increasing the batch size at each generation from $\sim10^2$ to $\sim10^5$, reduced the runtime of MAP-Elites by a few orders of magnitude ($\sim100\times$) without altering the performance, all on a single, customer-grade GPU.

However, naive random variations might not be the most effective method to utilise a large number of parallel evaluations.
For example, they remain inefficient and ineffective optimisers in domains that have high-dimensional search spaces.
Methods proposed to overcome this limitation include replacing or augmenting the random variations of vanilla QD algorithms with other optimisers, such as policy gradients (PG)~\citep{nilsson2021policy, pierrot2022diversity} or natural gradients in the form of Evolution Strategies (ES)~\citep{colas2020scaling, fontaine2020covariance}. 
ES-based QD approaches are demonstrated to be powerful and they can be applied more generally as they are black-box optimisers.
However, they require many samples to estimate gradients and become only practical and competitive to PG-based algorithms with access to large numbers of samples and evaluations~\cite{salimans2017evolution}.

Another case where random variations also struggle is in uncertain domains where naive single-sample estimates of solutions can be "lucky" and not reproducible, leading to a loss in true performance~\cite{adaptive, hbr, flageat2020fast, flageat2023uncertain}.
This problem setting is also referred to as Uncertain Quality-Diversity~\cite{flageat2023uncertain}.
Similarly, solutions to this problem commonly revolve around taking more samples of the same solution to obtain better estimates~\cite{hbr, adaptive} or other optimisation schemes which provide and demonstrate more implicit stability~\cite{flageat2020fast, flageat2022empirical}.

Both these cases are examples of settings where more effective methods than random variations exist but require a large number of evaluations, making them ideal settings for algorithmic innovation through the availability of large-scale parallel evaluations.
Ideally, we want methods that (1) are black-box optimisers so that they can be generally used across domains, (2) rely on efficient optimisers, to be effective in both low and high-dimensional domains, (3) obtain reproducible and reliable solutions in uncertain domains, and (4) can benefit and scale well with parallelisation, to leverage and maximise utilisation of more powerful hardware being developed.
It is important to note here that the use of the term scale does not just refer to the use of more compute or more devices. Instead, it also refers to better compute devices which allow for greater speed and parallelisation even on just a single easily-accessible and commonly available GPU device.

In this work, we address this by proposing \longname{} (\name{}), an ES-based QD algorithm that exploits large-scale parallelisation to improve the search for quality and diversity.
\name{} maintains multiple (up to $\sim100$) parallel and independent ES-processes each with its own objective, also called emitters, all on a single GPU device by leveraging tensorisation and parallel evaluations.
Exploration and exploitation objectives are distributed between the parallel ES emitters, which together simultaneously perform optimisation by generating new solutions.   

We compare the performance of \name{} against $10$ baseline QD algorithms and demonstrate that it outperforms them across 4 different tasks.
This includes beating state-of-the-art PG-based QD algorithms such as PGA-ME in high-dimensional and uncertain QD-RL neuroevolution tasks.
In uncertain domains, solution archives from \name{} prove as reproducible as purpose-built sampling-based approaches that handle uncertain settings well~\cite{flageat2020fast, flageat2023uncertain} while obtaining better performance, demonstrating the importance of effectively using samples and carefully considering them for addition.
Finally, we provide an extensive ablation of the main mechanisms of \name{}.
These results make \name{} a promising QD approach in settings where very large numbers ($\sim 10^5$) of evaluations are available at each generation.
Moreover, \name{} presents an approach which enables powerful and large-scale ES optimisation on just a single GPU.

\section{Background and Related Work}

\subsection{Quality-Diversity}

Quality-Diversity (QD)~\cite{cully2017quality} is an optimisation approach that seeks to find a collection of diverse and high-performing solutions to a problem.
In the QD setting, the diversity of solutions is defined according to a set of dimensions of interest referred to as features.
Thus, each solution has an attributed fitness $f$, quantifying its quality; and a feature $d$, quantifying its novelty with respect to other solutions. 
Most recent QD approaches rely on one of the two main QD algorithms: MAP-Elites~\cite{mouret2015illuminating} or Novelty-Search with Local Competition~\cite{lehman2011novelty}.
Our work builds upon MAP-Elites. 

\subsubsection{\textbf{MAP-Elites}} \label{subsec:ME-background}

MAP-Elites (ME)~\cite{mouret2015illuminating} is a common and popular QD algorithm.
ME keeps an archive $\archive{}$ of elite solutions encountered during optimisation.
To do so, it discretises the feature space into equally spaced grid cells, and keeps the highest fitness solution encountered in that feature cell during optimisation.
The algorithm follows the simple procedure of (i) selection from the archive $\archive{}$, (ii) update of selected solutions via perturbations (i.e. mutations), (iii) evaluation to obtain the corresponding fitness $f$ and feature $d$ (iv) update of archive  $\archive{}$ by attempting to add solutions.
A new solution is added to the archive  $\archive{}$ if it occupies a new cell (i.e. is novel) or if it is better performing than the solution in an occupied cell. This way the archive  $\archive{}$ incrementally improves as this optimisation procedure is repeated.
In this paper, we use the notion of emitter for ME introduced in~\cite{fontaine2020covariance}. An emitter is the generic name given to a procedure used to generate new offspring to be added to the archive $\archive$.
It corresponds to steps (i) and (ii) above.

\subsubsection{\textbf{Uncertain QD}} \label{subsec:uqd}
Uncertain QD (UQD) refers to the setting where the fitness and feature of solutions are no longer constants but instead are distributions denoted as $f \sim \mathcal{D}_f$ and $d \sim \mathcal{D}_d$~\cite{flageat2023uncertain, flageat2022empirical}.
In UQD domains, solutions can be "lucky" during an evaluation (i.e. outlier under the distribution) resulting in a fitness or features that make them appear more performing or more diverse. 
Due to their elitism, vanilla QD algorithms prioritise these "lucky" solutions over genuinely diverse or high-performing ones and thus return under-performing archives.
Thus, core challenges in the UQD setting are (i) performance estimation (ii) optimising for solutions that are reproducible~\cite {flageat2023uncertain}. 
(i) deals with correctly evaluating the true performance and novelty of a solution while (ii) tries to identify and prioritise solutions that exhibit consistent performance under uncertainty, also referred to as reproducible solutions. 
The most common UQD approach is ME-Sampling~\cite{hbr, adaptive, flageat2023uncertain}. It re-samples each solution $N$ times, and uses the average of the $N$ reevaluations to better estimate its fitness and feature before addition to the archive.
This helps to prevent illusory elitism.
Our approach implicitly helps with both UQD challenges without compromising on strong QD optimisation by using samples more effectively. 

\subsubsection{\textbf{Hardware-accelerated QD}}

Recent advances in hardware acceleration have led to new tools that run on devices such as GPUs and TPUs.
This can drastically reduce the evaluation time for optimisation algorithms. Examples of such tools are highly-parallel simulators like Brax~\cite{brax2021github} and Isaac~\cite{makoviychuk2021isaac}, which speed up robotic simulations.
These tools led to new QD libraries such as QDax \cite{lim2022accelerated}, that allow $10$ to $100$ times more evaluations per generation in parallel within the same amount of time.
Using these tools, Lim et al.~\cite{lim2022accelerated} demonstrated that the performances of ME are robust to large increases in the batch-size value. This drastically speeds up the run-time of QD algorithms by $\sim100\times$ and opens the door to promising future applications and algorithmic innovations.

\subsection{Evolution Strategies} \label{subsec:ES}

Evolution Strategies (ES) are black-box optimisation algorithms inspired by evolution where a parameterised distribution over solutions is updated in the direction of higher fitness solutions~\cite{ beyer2002evolution, wierstra2014natural}. 
The different ES are classified according to their sampling and update procedure of the parameterised distribution.
We focus on the ES introduced by Salimans et al.~\cite{salimans2017evolution} for its ability to operate effectively on high-dimensional RL tasks. This algorithm is also commonly referred to as OpenAI-ES but we refer to it as ES for brevity from this point onwards.
ES uses the natural gradient approximated from samples to update the parameters of the search distribution. 
The distribution is represented as an isotropic multivariate Gaussian distribution $\mathcal{N}(\theta_\mu, \sigma^2I)$, where $\theta_\mu$ and $\sigma^2$ are the mean and variance of the distribution. In this ES variant, $\sigma$ remains constant and is a fixed hyperparameter. Therefore, the distribution is only parameterised by its mean value $\theta_\mu$.
A batch or population of $N$ search points are sampled from this parameterised distribution $\theta_i = \theta_\mu + \sigma\epsilon_i$ where $\epsilon_i \sim \mathcal{N}(0, I)$.
Each sampled point $\theta_i$ is then evaluated on the objective function $F$ and used to estimate the search gradient to update the parameters of the distribution:
\begin{align}
    \nabla E_{\theta\sim\mathcal{N}(\theta_\mu, \sigma^2I)}[F(\theta)] \approx \frac{1}{N\sigma}\sum{F(\theta_i)\epsilon_i}
\end{align}

We give the pseudo-code for one ES optimisation step in Appendix~\ref{app:exp_details}.
This procedure is then repeated iteratively to improve the parameters of the search distribution.
Following Salimans et al.~\cite{salimans2017evolution}, we also perform fitness shaping in the form of rank normalisation.

\subsubsection{\textbf{Novelty-Search with Evolution Strategies}} \label{subsec:NS-ES}

The variant Novelty-Search ES (NS-ES)~\cite{conti2018improving} also uses ES, described in the previous section, but augments the objective term to be able to handle sparse and deceptive rewards to encourage exploration in such settings. 
To do so, NS-ES uses a novelty score as the objective $F$ for the ES. 
Taken from Novelty Search~\cite{lehman2011evolving}, the novelty score of a solution $\theta$ corresponds to the average distance in feature $d$ to its K-nearest neighbours in a novelty archive $\noveltyarchive{}$, containing all previously evaluated solutions: 
$
    novelty(\theta) = \frac{1}{K}\sum_{k=1}^{K} \|d_{\theta} - d_k \|_2
$.

\subsection{Gradient-augmented QD} \label{subsec:gradient_QD}

Despite the simplicity and success of conventional QD algorithms, they have generally been shown to be efficient in low-dimensional search spaces but suffer in high-dimensional ones due to the sample-inefficiency of their GA-based mutation~\cite{colas2020scaling}.
To be able to function in applications such as neuroevolution, where the size of the parameter space is easily in the thousands, more efficient optimisation procedures were needed over the standard GA mutations used, such as gradient-based optimisation methods, that are generally known to be more efficient optimisation algorithms.

Our work builds on MAP-Elites ES (ME-ES)~\cite{colas2020scaling}, detailed in Section~\ref{subsec:ME-ES}, which uses ES and maintains an archive of elites. 
Other QD algorithms that use ES also exist such as Novelty Search-Reward ES (NSR-ES) and Novelty Search-Reward Adaptive ES (NSRA-ES)~\cite{conti2018improving} which also uses ES~\cite{salimans2017evolution} but instead optimises for a weighted sum of the task reward and the novelty score as the objective.
Covariance Matrix Adaptation ES (CMA-ES)~\cite{hansen2006cma}, an alternative ES approach, has also been combined with ME resulting in CMA-ME~\cite{fontaine2020covariance}.
The space and computational time complexity prevents CMA-ES and hence CMA-ME from being used on high-dimensional search spaces.
However, recent extensions~\cite{tjanaka2022approximating, tjanaka2022training} try to address this challenge by approximating the covariance matrix.

Another class of gradient-based QD methods tailored to Markov Decision Processes (MDP) and Deep Reinforcement Learning (DRL) problems~\cite{arulkumaran2017deep} are actor-critic based methods such as Policy Gradient Assisted ME (PGA-ME)~\cite{nilsson2021policy} and QD-PG~\cite{pierrot2022diversity}. 
They utilize policy-gradients (PG) from DRL which rely on training a critic network to apply variations to parent policies.
They have shown to be effective in the QD-RL setting but require training an additional critic network.
In the following, we refer to QD approaches using NES mutations as ES-based QD. Similarly, we refer to approaches mixing GA and PG as PG-based QD.

\subsubsection{\textbf{MAP-Elites with Evolution Strategies}} \label{subsec:ME-ES}

MAP-Elites ES (ME-ES)~\cite{colas2020scaling} combines the optimisation procedure of ES~\cite{salimans2017evolution} with the maintenance of an archive in ME~\cite{mouret2015illuminating}.
It uses the ES optimisation procedure in two separate sequential modes; "exploit" and "explore".
The "exploit" mode is used with the task fitness as the objective $F$ for the ES while the "explore" mode uses a novelty score as the objective $F$.
For the "explore" mode, ME-ES maintains a novelty archive $\noveltyarchive{}$ (see Section~\ref{subsec:NS-ES}), separate and different from the archive of elites $\archive{}$.
ME-ES starts by selecting a solution from the archive of elites $\archive{}$ and applies one of the modes (e.g. "exploit") for $\optim{}$ generations. 
Each generation corresponds to one gradient step (update of the distribution). 
At every generation, the updated mean of the distribution is considered for addition to the archive $\archive{}$. 
After $\optim{}$ generations, ME-ES selects another solution from the archive and applies the other mode (e.g. "explore").
ME-ES alternates between the "exploit" and "explore" modes, stopping each mode after $\optim{}$ generations (in the original ME-ES, $\optim{} = 10$).

After stopping a mode, a new solution is selected from the archive $\archive{}$ following a biased selection procedure rather than a random uniform selection commonly used in ME.
This procedure aims to select solutions that are already high-performing and can be improved further. It selects either one of the highest performing for exploit mode or one of the most novel solutions for explore mode.  
The pseudo-code for ME-ES is provided in Appendix~\ref{app:exp_details}.

\section{Methods}

In this paper, we introduce \longname{} (\name{}), which leverages tensorisation and advances in hardware accelerated QD to maintain multiple independent ES workers in parallel each with its own independent objective
and reset mechanism designed for QD optimisation. 
In the following we use $32$ parallel ES but we show in Appendix~\ref{app:more_results} that MEMES also scales to $128$ parallel ES. 
These modifications address commonly found limitations of the ME-ES algorithm (see Sec.~\ref{subsec:ME-ES}) such as slow and inefficient optimisation due to the large number of samples and low batch-size.

\subsection{Parallel ES: from sequential to simultaneous}

\name{} leverages the use of vectorisation and fast parallel evaluations~\cite{brax2021github} to be able to run hundreds of independent ES in parallel.
Each independent ES represents a separate emitter that generates new solutions. 
In contrast, in ME-ES, a single thread of ES at each generation sequentially alternates between the exploitation of the task fitness objective and of the novelty objective (see Section~\ref{subsec:ME-ES}).
To effectively do this, we reduce the number of samples $\nsamples{}$ (i.e. population size) used by each ES process to estimate the gradient. 
In ME-ES~\citep{colas2020scaling}, a single ES update (i.e. gradient step) uses $\nsamples{}=10,000$ samples. 
Here, we use $\nsamples{}=512$ samples per ES emitter as we observe an insignificant increase in performance when using larger values of $\nsamples{}$ beyond $512$ in many of our experiments. 
Thus, for comparison, to generate $32$ offspring (the value used in this work), \name{} uses $16384$ samples per generation while ME-ES uses $10000$ samples to generate $1$ offspring.

This parallelisation allows multiple independent emitters to be run together in order to perform the explore and exploit modes simultaneously. 
In \name{}, a proportion $p_{exploit}$ of the emitters is dedicated to the exploitation of the task fitness objective, and a proportion $1 - p_{exploit}$ is dedicated to the optimisation of the novelty score.
While our experiments only use these objectives for comparison, other objectives such as improvement~\citep{fontaine2020covariance}, curiosity~\citep{pathak2017curiosity, lim2022learning} and others can also be used at the same time~\citep{cully2021multi}.
This enables many objectives to be optimised simultaneously and in parallel.
This interplay could lead to the generation of new and interesting solutions and greater efficiency and more effective search.

\subsection{Dynamic emitter reset}

\SetCommentSty{mycommfont}
\begin{algorithm}[t]
\footnotesize
\DontPrintSemicolon
\SetKwBlock{DoParallel}{do in parallel for}{end}
\caption{\name{} proposed in this work.} 
\label{alg:our-es}
    \textbf{Inputs:} sample size $N$, number of generations $N_{gen}$, number of emitters $N_{emitter}$, proportion of exploit emitter $p_{exploit}$

    \textbf{Initialisation:} archive of elites $\archive{}$ and novelty-archive $\noveltyarchive{}$
    
    \tcp{Set objective for each emitter}
    $n = N_{emitter} * p_{exploit}$ \;
    \lForEach{$e = 1, ... n$}{
        $F_e \gets \text{fitness\_objective}()$
    }
    \lForEach{$e = n, ... N_{emitter}$}{
        $F_e \gets \text{novelty\_objective}(\noveltyarchive{})$
    }

    \tcp{Main loop}
    \For{$g = 1, ... N_{gen}$}{

        \DoParallel($e = 1, ... N_{emitter}$){

            \uIf{require\_reset}{
            
                $\theta \gets \text{uniform\_selection}(\archive{})$ \;
                $\stagnate{} = 0$ \;
                $\text{require\_reset} = False$ \;
            }
            
            $\tilde\theta \gets \text{ES\_step}(F_e, N, \theta) $ \;
            $\archive{}, \noveltyarchive{} \gets \text{update\_archives}(\tilde\theta)$ \;
            \uIf{$\text{added\_to\_archive}(\archive{}, \tilde\theta)$}{
                $\stagnate{} = 0$ \;
            }
            \Else{
                $\stagnate{} = \stagnate{} + 1$ \;
                \lIf{$\stagnate{} > \stagnatebudget{}$}{
                    $\text{require\_reset} = True$}
                }
            
        }
    }
\Return $\archive{}$
\end{algorithm}
\setlength{\textfloatsep}{5pt}

An ES-emitter reset refers to selecting a new solution to restart an ES optimisation process from. This is done by selecting a solution from the ME archive $\archive{}$.
In ME-ES, the ES-emitter is reset every fixed number of generation $\optim{}$ to switch between explore and exploit mode. 
This is imposed by the constraint of having only a single emitter that needs to switch between the two modes. 
However, this might impair the optimisation process as solutions with high evolvability, dwelling in promising areas of the search space, might not be fully exploited. 
Additionally, as the optimiser state is reset with the emitter itself, even if an evolvable solution is later re-sampled, it would still take a few generations to be able to use it meaningfully.
Our experimental results (shown in the following sections) indicate that some emitters can generate promising individuals for hundreds of generations, beyond the optimisation lifetime imposed in ME-ES ($\optim{} = 10$).

\name{} introduces a reset strategy where each independent emitter can be reset separately to fully maximise having simultaneous explore and exploit emitters in parallel.
We propose dynamically and automatically resetting the exploit and explore emitters based on their usefulness to the QD optimisation. 
Each ES emitter generates one offspring $\tilde{\theta}$ per generation (gradient step), which is a candidate to be added to the archive $\archive{}$. Each time a candidate is not added to $\archive{}$, we increase a \stagnatename{} $\stagnate{}$; each time the candidate is added to $\archive{}$, we reset $\stagnate{}$ to $0$.
This introduces a new hyperparameter: the \stagnatebudgetname{} $\stagnatebudget{}$.
If $\stagnate{}$ exceed $\stagnatebudget{}$, meaning if the emitter does not generate promising offspring for more than $\stagnatebudget{}$ consecutive generations, the emitter is reset.
This mechanism allows \name{} to fully leverage lineages of promising solutions, and automatically determine good optimisation paths.
We demonstrate the benefits of this mechanism in an ablation study in the experimental section. 

When resetting an ES emitter, a random solution is selected with a uniform probability from the archive as a new starting point of the ES process.
This is done to reduce bias on any particular task and increase the probability of utilising stepping stones from unexpected feature space regions~\cite{mouret2015illuminating}. 
More details regarding this is available in Appendix~\ref{app:exp_details}.

\subsection{Improving the explore emitter} \label{subsec:methods_explore_emitter}

Different QD algorithms use different strategies for exploration. 
The most commonly used strategy is random mutations (GAs)~\cite{mouret2015illuminating, nilsson2021policy}.
However, as motivated, solutions from these operators might not be the most effective way to leverage access to a large number of evaluations.
Other approaches that rely on explicit novelty optimisation also exist.
QD-PG \cite{pierrot2022diversity}, which only applies in RL-MDP settings, follows a state-novelty gradient. CMA-MEGA \cite{fontaine2021differentiable} follows feature gradients in whitebox optimisation settings, but it can also be applied to problems where gradients can be approximated \cite{tjanaka2022approximating}.
In \name{}, following NS-ES, NSR-ES~\citep{conti2018improving} and ME-ES~\citep{colas2020scaling}, we compute approximated-gradients using ES to move in the direction of high novelty-score. 
As explained in Section~\ref{subsec:NS-ES}, computing the novelty score requires a novelty-archive $\noveltyarchive{}$, consisting of all previously evaluated solutions, regardless of how they perform.
Note that this is separate and different to the archive of elites $\archive{}$, which contains only the best-performing solutions (see Section~\ref{subsec:NS-ES}).
Maintaining this novelty archive $\noveltyarchive{}$ is extremely memory-intensive, especially in highly-parallel settings, and requires infinitely increasing computation as the archive expands linearly with generations.
Hence, in \name{}, we propose and use a first-in-first-out (fifo) novelty archive to overcome this challenge.
The fifo novelty archive stores all evaluated offspring until it reaches full capacity. Then, each time a new solution enters the fifo novelty archive, it replaces the oldest solution.
We also explore alternative approaches to compute novelty scores but demonstrate experimentally that this simple approach works the most effectively while alleviating memory usage.
We consider the following approaches:

\begin{itemize}[leftmargin=*]
    \item \textbf{Novelty-archive-fifo-ES explore (Ours):} uses the fifo novelty archive described before which alleviates memory congestion.
    \item \textbf{Novelty-archive-all-ES explore:} conventional novelty archive $\noveltyarchive{}$ that keeps all evaluated solutions. This has computational time and memory limitations but is used as a baseline.
    \item \textbf{Elites-archive-ES explore:} replaces the novelty archive with the QD archive of elites $\archive{}$, when computing novelty score.
    This requires significantly less memory as only the elite of each niche are kept in $\archive{}$.
    However, this also leads to a loss of historical and cumulative information about solutions seen before.
    The QD archive only considers the filling of an empty cell as novel while a novelty archive keeps all solutions seen during search, and accounts for the density of solutions in an area of the feature space when computing novelty.
    This means that a QD-archive cannot identify when a part of the feature space has been explored many times unsuccessfully.
    
    \item \textbf{GA:} uses random mutation~\cite{vassiliades2018iso}. Non-archival method which does not require any memory resources at all. 
\end{itemize}

\section{Experiments}

\subsection{Experimental Setup} \label{subsec:exp-setup}

\subsubsection{Environments}

We evaluate \name{} and baselines on four tasks from QD literature. Two of these tasks (Arm and Hexapod) are black-box deterministic QD benchmarks while the other two tasks (Ant and AntTrap) are neuroevolution RL tasks in the UQD setting. Both neuroevolution tasks consider noise in the initial state and in the dynamics of the rollout induced by the simulator.

\begin{itemize}[leftmargin=*]
    \item \textbf{Redundant Arm (Arm)} is a commonly used as a benchmark QD task~\cite{cully2015robots, cully2017quality, gaier2020discovering, fontaine2021differentiable}. This task aims to learn the smoothest angular-position vectors to reach all $(x, y)$ positions with a $1000$-DoF planar robotic arm.

    \item \textbf{Omnidirectional Hexapod (Hexapod)} is a QD benchmark that aims to learn the vectors of a parameterised open-loop controller to walk in every direction with a 18-DoF Hexapod, while minimising its orientation error~\cite{cully2013behavioral, chatzilygeroudis2018reset}.

    \item \textbf{Unidirectional Ant (Ant)} proposed as part of the QDGym suite \cite{nilsson2021policy, flageat2022benchmarking}. In this neuroevolution task, the algorithms learns closed loop neural network  policies for a four-legged Ant robot with randomised initial state. The objective is to find a set of diverse gaits for moving forward as fast as possible.
    
    \item \textbf{Ant Trap} is a neuroevolution task similar to Ant but consists of deceptive rewards and is categorized as hard-exploration~\cite{pierrot2022diversity, chalumeau2022assessing}.
    In Ant Trap, exploiting the task fitness directly would result in getting stuck in the trap present in the environment. 
    In contrast, exploitation of the task fitness in the Ant task is always a good idea.
    Additionally, in Ant Trap, the policies that are close to the origin will inherently have a low task fitness and not solve the task (i.e. pass the trap) despite contributing to the diversity.
    To the best of our knowledge, this task has not been considered and evaluated in the UQD setting prior to this work, making it an interesting task. 
\end{itemize}
Detailed descriptions of the fitness and feature (including equations) for each task is provided in Appendix~\ref{app:exp_details}.

\subsubsection{Baselines}

In the following, we compare \name{} to a wide range of baselines. We also consider a variant of our approach named \name{}-all, which considers all samples and not just offspring of ES emitters for addition to the archive. 
\begin{itemize}[leftmargin=*]
    \item \textbf{ES}~\cite{salimans2017evolution} from Sec.~\ref{subsec:ES} optimises for task fitness.
    \item \textbf{NS-ES}~\cite{conti2018improving} from Sec.~\ref{subsec:NS-ES} optimises for novelty score.
    \item \textbf{NSR-ES and NSRA-ES}~\cite{conti2018improving} from Sec.~\ref{subsec:gradient_QD}. They use a weighted average of the task fitness and the novelty score as objective. 
    \item \textbf{ME}~\cite{mouret2015illuminating} which uses random mutations (GAs). Here, we use the line mutation operator~\cite{vassiliades2018iso}. We consider two variants of ME. ME-128 with a batch size of 128 which is a commonly used value and ME-16384 with a batch size of 16384 which corresponds to the same number of samples used by \name{} at each generation.
    \item \textbf{ME-Sampling}~\cite{hbr} Given that re-evaluation is one of the most common methods in uncertain domains~\citep{flageat2023uncertain}, we also consider the ME-sampling method which uses the same $16384$ evaluations at each generation but distributed to re-evaluate solutions. Here, each of the solutions is re-evaluated $32$ times~\cite{flageat2023empirical} to obtain a better estimate of $f$ and $d$ before addition.
    \item \textbf{ME-ES}~\cite{colas2020scaling} from Sec.~\ref{subsec:ME-ES} alternating explore and exploit ES.
    \item \textbf{PGA-ME}~\cite{nilsson2021policy} described in Sec.~\ref{subsec:gradient_QD}. applies policy gradient (PG) updates to half of the selected solutions, in addition to GAs. 
    PGA-ME represents the upper baseline and state-of-the-art for high-dimensional neuroevolution QD-RL tasks such as Ant and AntTrap. However, it can only be used for RL tasks with MDP assumptions.
    \item \textbf{CMA-ME}~\cite{fontaine2020covariance} described in Sec.~\ref{subsec:gradient_QD}.
    CMA-ME is only used as a baseline for the lower-dimensional optimisation task (i.e. non-neuroevolution). We do not compare to its extension CMA-MEGA with approximated gradients as it demonstrates similar or lower performing results than PGA-ME on QD-RL tasks~\cite{tjanaka2022approximating}.
\end{itemize}

\begin{figure*}[t!]
\centering
\includegraphics[width = 0.95\hsize]{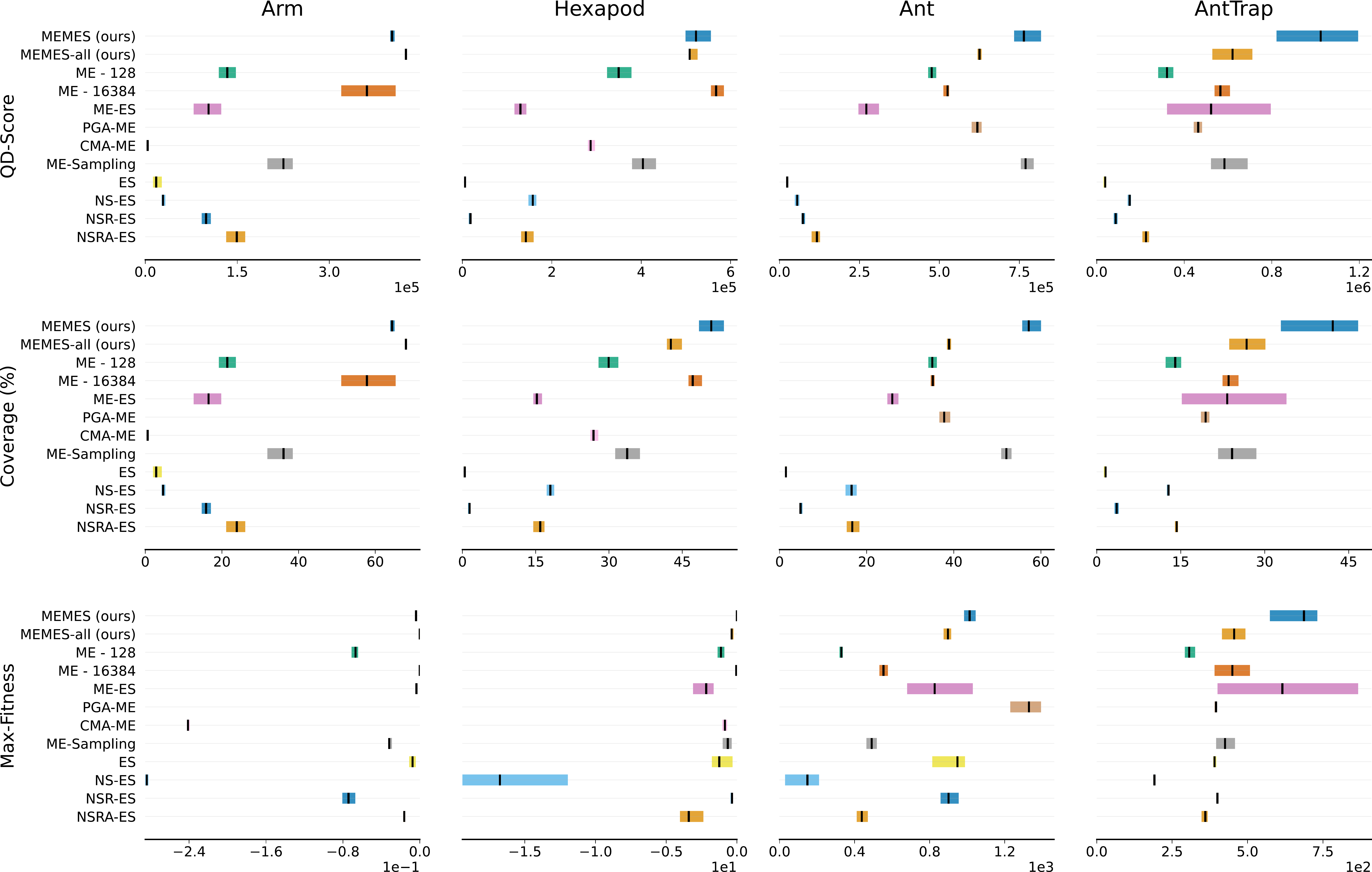}
\caption{
    Final QD-Score (top), Coverage (middle) and Max-Fitness (bottom). 
    We report the median and CI over $10$ seeds. 
}
\label{fig:main_result}
\end{figure*}

It is important to note that ES, NS-ES, NSR-ES and NSRA-ES do not have an archive of elites and do not rely on re-selection from the archive during the entire optimisation process. 
For comparison, we maintain a passive archive of elites and consider the addition of the mean of the ES at every generation.
We open-source our code at~\url{https://github.com/adaptive-intelligent-robotics/MEMES}.

\subsubsection{Metrics}

We consider the QD-Score, Max-Fitness, and Coverage metrics~\cite{pugh2016quality} when conducting our analysis.
In our evaluation, all three metrics provide valuable insights: the effect of explore and exploit emitters can be respectively assessed from the Coverage and Maximum-Fitness scores, while the QD-score evaluates the overall algorithm performance. 
In the uncertain domains (neuroevolution tasks), we consider the Corrected-metrics~\citep{adaptive, flageat2023uncertain}, computed by re-evaluating each solution in the archive $512$ times and adding them to an empty archive.
We run Arm, Hexapod and Ant for $2000$ generations and AntTrap for $10000$ generations. 
We compute p-values based on the Wilcoxon rank-sum test with Bonferroni correction.

\subsection{Overall Performance}

Figure~\ref{fig:main_result} shows the performance of \name{} compared to the baselines.
\name{} performs better than all baselines on Arm ($p < 0.08$ for QD-Score and Coverage).
On the Hexapod, \name{} performs on par with ME-16384 while outperforming all other baselines ($p < 0.03$).
Additionally, the MEMES-all variant outperforms MEMES on Arm ($p < 2.10^{-4}$ for QD-Score and Coverage), where the attempted addition of more samples (from the ES) to the archive results in better performance, but performs on par with \name{} on Hexapod.
In the Ant neuroevolution task, we observe that \name{} performs better than PGA-ME, which is known to consistently do well on this task~\citep{nilsson2021policy, tjanaka2022approximating}, in both QD-Score and Coverage ($p < 0.003$) but reaches a lower Max-Fitness.
The Max-Fitness of \name{} seems to be constrained by the performance of the ES algorithm as we observe Max-Fitness performance converging to that of conventional ES.
This makes sense given that the adaptive emitter reset in \name{} would enable a conventional ES optimisation thread to be present.
However, the benefit of parallel explore and exploit ES emitters each following independent gradients in \name{} is still overall evident from the higher Coverage compared to GA-based approaches such as ME and PGA-ME ($p < 0.003$ for Coverage) and also single ES-based approaches such as ES, NS-ES and NSR-ES.
In the AntTrap neuroevolution task, \name{} successfully performs the best across all baselines, proving the only approach to solve the task and move around the deceptive trap. 
Recall that an uncertain version of these neuroevolution tasks are considered and the corrected metrics are used in the comparison.


These results suggest that with access to cheap and large numbers of evaluations, \name{} can be the go-to option across a range of domains. 
Given the use of vectorisation and parallelisation of evaluations, powerful large-scale search optimisation (\name{}) is accessible and practical to be run in a reasonable time-frame with only a marginal increase in run-time.
An analysis of the run-time of \name{} and baselines are provided in Appendix~\ref{app:more_results}.
Resulting archives of elites on the various tasks can be found in the Appendix~\ref{app:more_results} and videos of resulting behaviours are available at: \url{https://sites.google.com/view/memes-qd/home}.

\begin{figure}[t!]
\centering
\includegraphics[width = \hsize]{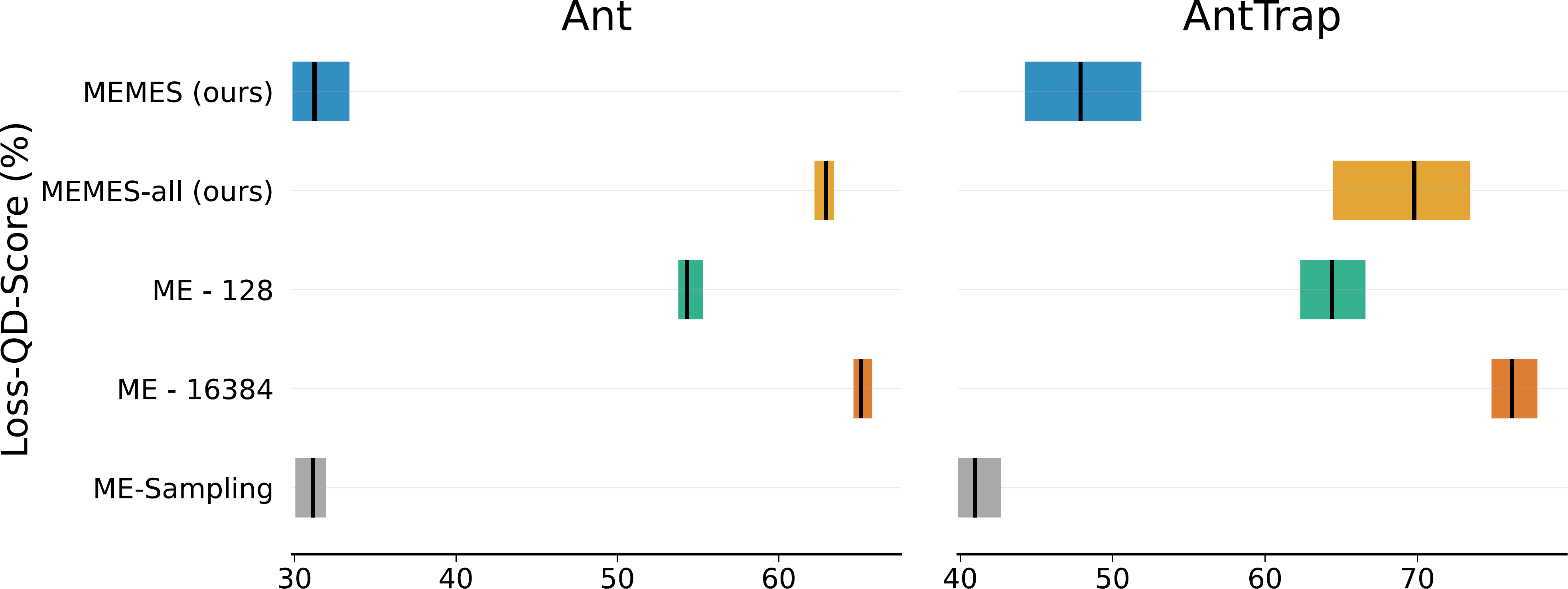}
\caption{
    QD-Score loss that quantifies the ability of algorithms to correctly estimate the performance of solutions in UQD setting. We report the median and CI over $10$ seeds.
}
\label{fig:archive_loss}
\end{figure}

\subsubsection{Sampling-wise Comparison and Reproducibility}

\name{} requires a large number of evaluations to approximate gradients. 
We evaluate if this could be effective in handling the UQD problem (see Sec.~\ref{subsec:uqd}).
On the Ant and AntTrap tasks, we observe in Figure~\ref{fig:main_result} that samples used by \name{} are indeed more effective as they result in higher QD-scores than those of ME-16384 and also of ME-Sampling and ME-16384 for AntTrap.

\begin{table}[]
\footnotesize
\centering
\begin{tabular}{l|cccc}
\toprule

 & \textsc{Arm}
 & \textsc{Hexapod}
 & \textsc{Ant}
 & \textsc{AntTrap}
\\ 

\midrule
\addlinespace

\textsc{Explore emitter}
& 143.0
& 362.4
& 249.5
& 270.1
\\

\textsc{Exploit emitter}
& 265.7
& 337.2
& 366.5
& 387.7
\\

\bottomrule
\end{tabular}
\caption{
    Average lifespan (in num. of generations) of exploit and explore emitters when using adaptive reset in MEMES.
}
\label{tab:reset}
\vspace{-3mm}
\end{table}

On top of the corrected-metrics, we also further analyse the archive-reproducibility, by evaluating the algorithms on the archive loss metric. 
The archive loss is defined as the percentage difference before and after computing the corrected archive~\citep{flageat2022empirical}.
In Figure~\ref{fig:archive_loss}, \name{} is shown to have a much lower QD-Score loss than the ME baselines. \name{} losses are comparable to those of ME-Sampling, which is considered a strong baseline for this metric~\cite{flageat2022empirical}, while ME-Sampling has a significantly lower QD-Score on AntTrap. 
This demonstrates that \name{} archives can reliably maintain its quality and diversity and that \name{} better estimates and reflects the true performance in its metrics.
This ability of \name{} to favour solutions that reproducibly solve the task likely leads to it outperforming other approaches on AntTrap, while ME-16384 suffer from elitism and archive loss even with a significant number of iterations, and ME-sampling is crippled from inefficient optimisation.
\name{}-all also suffers from a high archive loss as it also considers many "lucky" solutions (from ES samples) for addition causing illusory archive elitism as well.
This means that only the offspring of ES emitters are more reproducible and further highlights the importance of carefully and effectively using samples and considering them for addition in uncertain domains.

\begin{figure}[t!]
\centering
\includegraphics[width = \hsize]{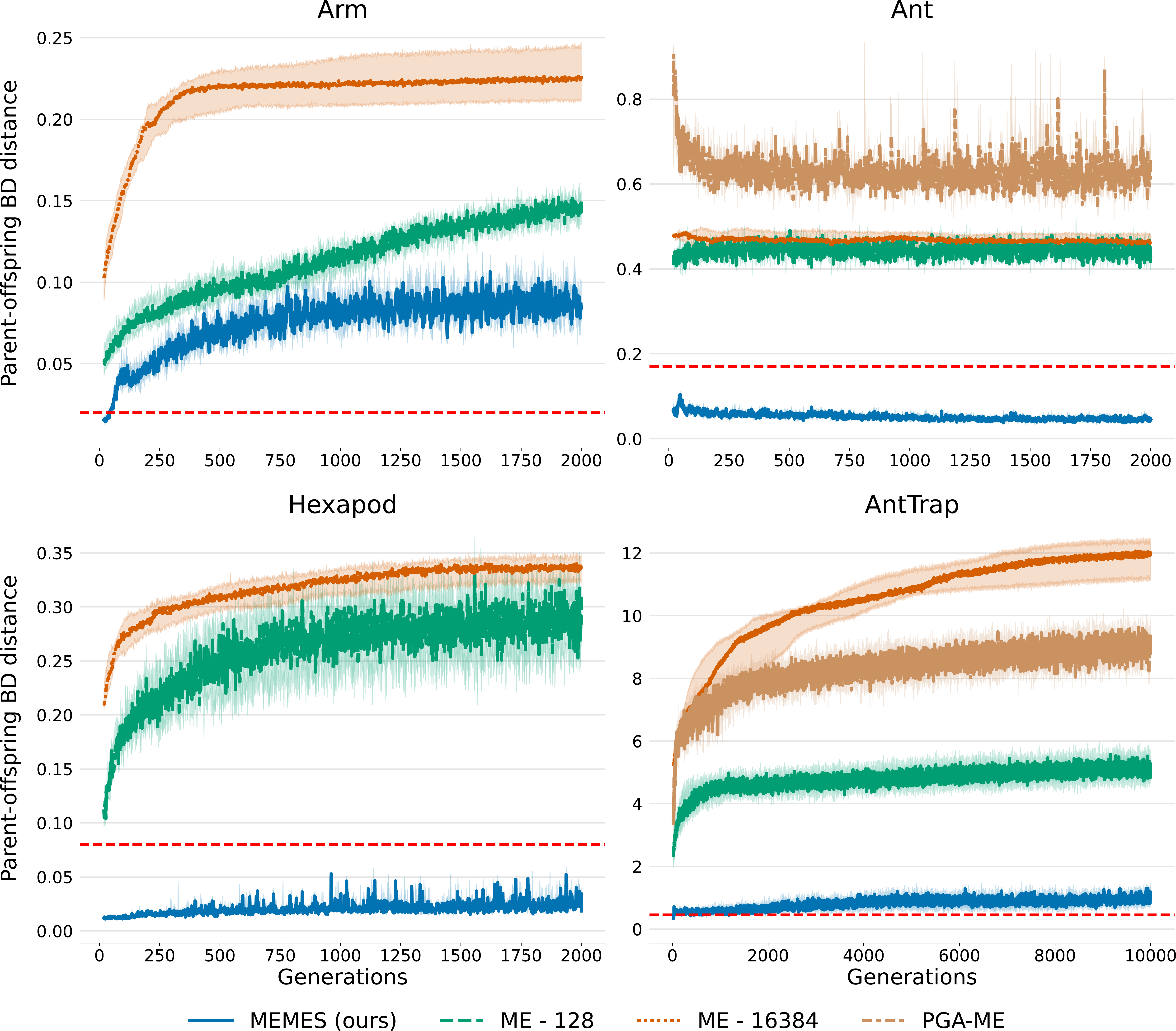}
\caption{
    Parent-offspring feature-distance for the exploit-ES emitter of \name{}, the PG emitter of PGA-ME and the GA emitter of ME. 
    The solid line is the median and the shaded areas are the quartiles over $10$ replications.
    We display in red the average width of the cells in the each task.
}
\label{fig:local_optim}
\end{figure}

\subsubsection{Effective Local Optimisation}

We also evaluate some characteristics of the optimisation paths and behaviour of \name{}.
Interestingly, we find in \name{} the ability to perform optimisation locally within or surrounding a cell niche, which is uncommon in QD algorithms.
This is an important characteristic in QD as optimising locally in low-performing niches/cells is crucial to ensure high fitness across the entire feature space.
This ability was measured using the distance between each parent and its offspring in the feature space.
This metric measures whether generated individuals stay within or around the vicinity of the same cell.
As the exploration aspect of QD algorithms such as novelty-ES are expected to maximise the distance in feature space, we only evaluate this metric for the task-fitness-maximising aspect of algorithms. 
For \name{}, we compute it for the exploit-ES and compare it to the PG variation of PGA-ME, and GA variation of ME.

Figure~\ref{fig:local_optim} shows this distance measure plotted across generations of the algorithm.
For comparison, we display on the same graph the cell-size for each task (red line).
In Ant and Hexapod, the exploit-ES emitters produce offspring solutions after a gradient step that have a distance of less than one cell (below red line). 
This demonstrates some implicit ability of the emitter to perform local optimisation around a certain niche. 
This is beneficial in tasks where the task fitness is orthogonal to the feature.
In the AntTrap tasks, the goal of the exploit-ES emitter is to move forward as fast as possible and hence the nature of this task does not require such local optimisation.
Interestingly, we also observe that the exploit-ES emitter largely results in offspring with much lower distance than PGA-ME and ME.
PGA-ME uses policy gradient variations that rely on training a global critic.
We hypothesise that this global critic systematically pushes updates in the same direction of the search space, which while can improve the policy might also result in similar features far away from parent solutions.
Likewise, the GA variation of ME has no specific mechanism to optimise solutions while remaining locally in a feature region and hence results in offspring further away in the feature space.


\begin{figure}[t!]
\centering
\includegraphics[width = \hsize]{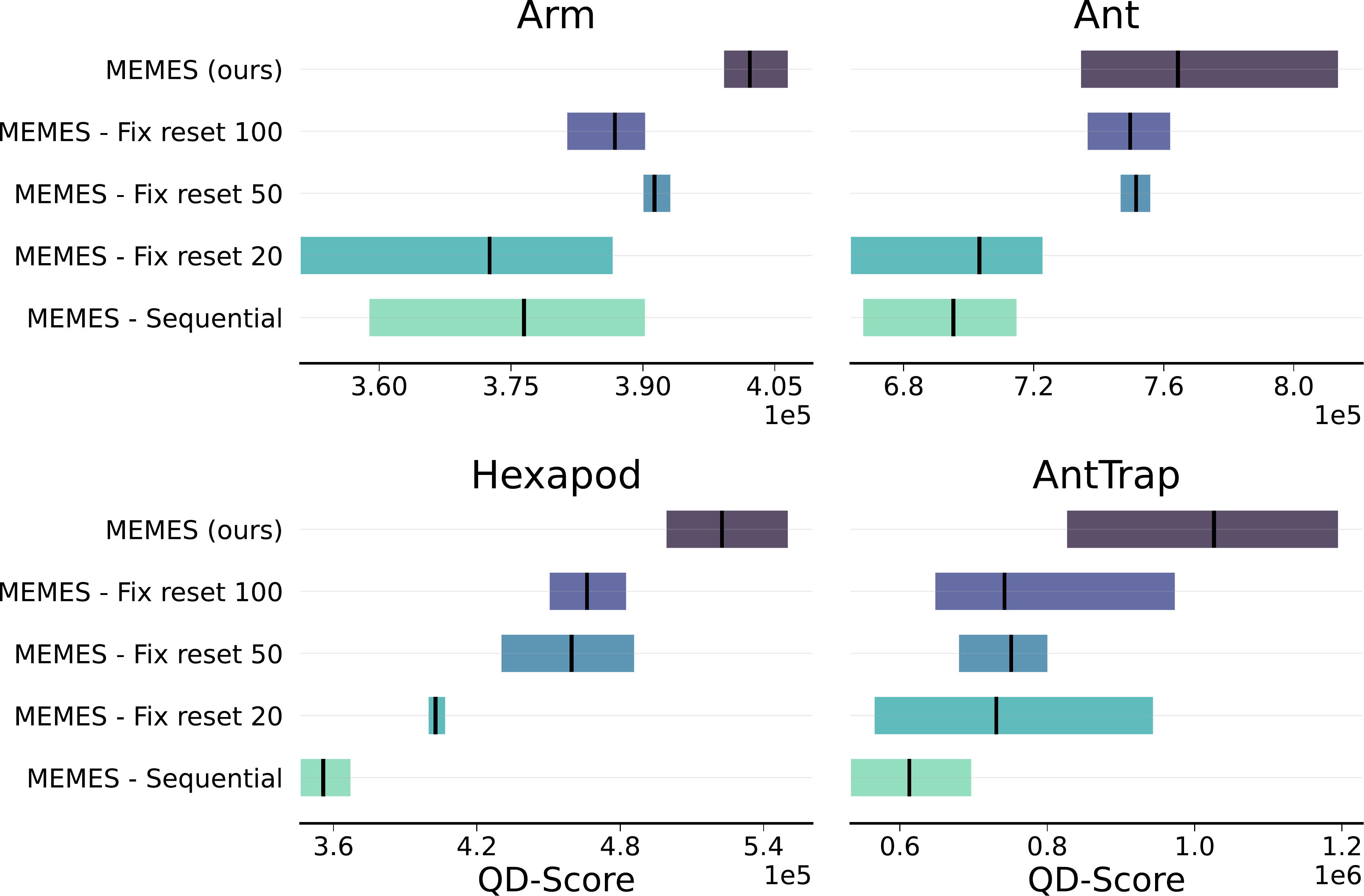}
\caption{
    Comparison of the final QD-score for different fixed reset values, compared to the adaptive reset mechanism. We report the median and CI over $5$ seeds. 
}
\label{fig:reset_ablation}
\end{figure}

\subsection{Analysis and ablation of \name{}}
In this section, we conduct experiments to study the different components of \name{} and show that each one of them contribute to the effectiveness of the approach.

\subsubsection{Simultaneous over sequential.}
To demonstrate the importance of having simultaneous explore and exploit emitters, we compare against \name{}-Sequential, a variant of \name{} with parallel ES emitters that follows the original ME-ES implementation of alternating between either fully explore or fully exploit modes. The emitters are reset and swapped mode after 10 generations.
It is important to note that being sequential necessitates having a fixed number of generations in order to swap between modes.
Figure~\ref{fig:reset_ablation} shows that \name-Sequential slightly under-performs \name{} with simultaneous explore and exploit (\name{} - Fix reset 10).
While the improvement of having simultaneous emitters is marginal, they allow independent ES emitters with adaptive resets.

\subsubsection{Adaptive emitter resets.}
Next, we evaluate the importance of the adaptive reset of the emitters by increasing the number of optimisation steps of ES before resetting the emitter.
We observe in Figure~\ref{fig:reset_ablation} that increasing the number of optimisation steps of the ES emitters seems to help (i.e. from $20$ to $100$). 
However, the optimal value may vary depending on the tasks.
For example, we can observe that there are minimal gains (Hexapod, Ant, AntTrap) or detrimental to overall performance (Arm) when moving from $50$ to $100$ optimisation steps.
Similarly, the optimal value also could depend on the mode of ES used (exploit or explore).
On the Ant task, more steps are critical for the exploit-ES for optimal maximum fitness performance while potentially unnecessary for the explore-ES model (coverage metric).
Conversely, in the AntTrap, more steps are important for the explore-ES (coverage) while the gain in steps might even be detrimental to exploit-ES at times.
Table~\ref{tab:reset} shows that the adaptive reset mechanism in \name{} can modulate and shows different average lifespans of the ES emitters depending on the task and the mode of the ES.
Figure~\ref{fig:reset_ablation} demonstrates that our adaptive mechanism achieves the best on all the metrics across all the tasks without much hyperparameter tuning.

\subsubsection{Explore Emitter.}
\begin{figure}[t!]
\centering
\includegraphics[width = \hsize]{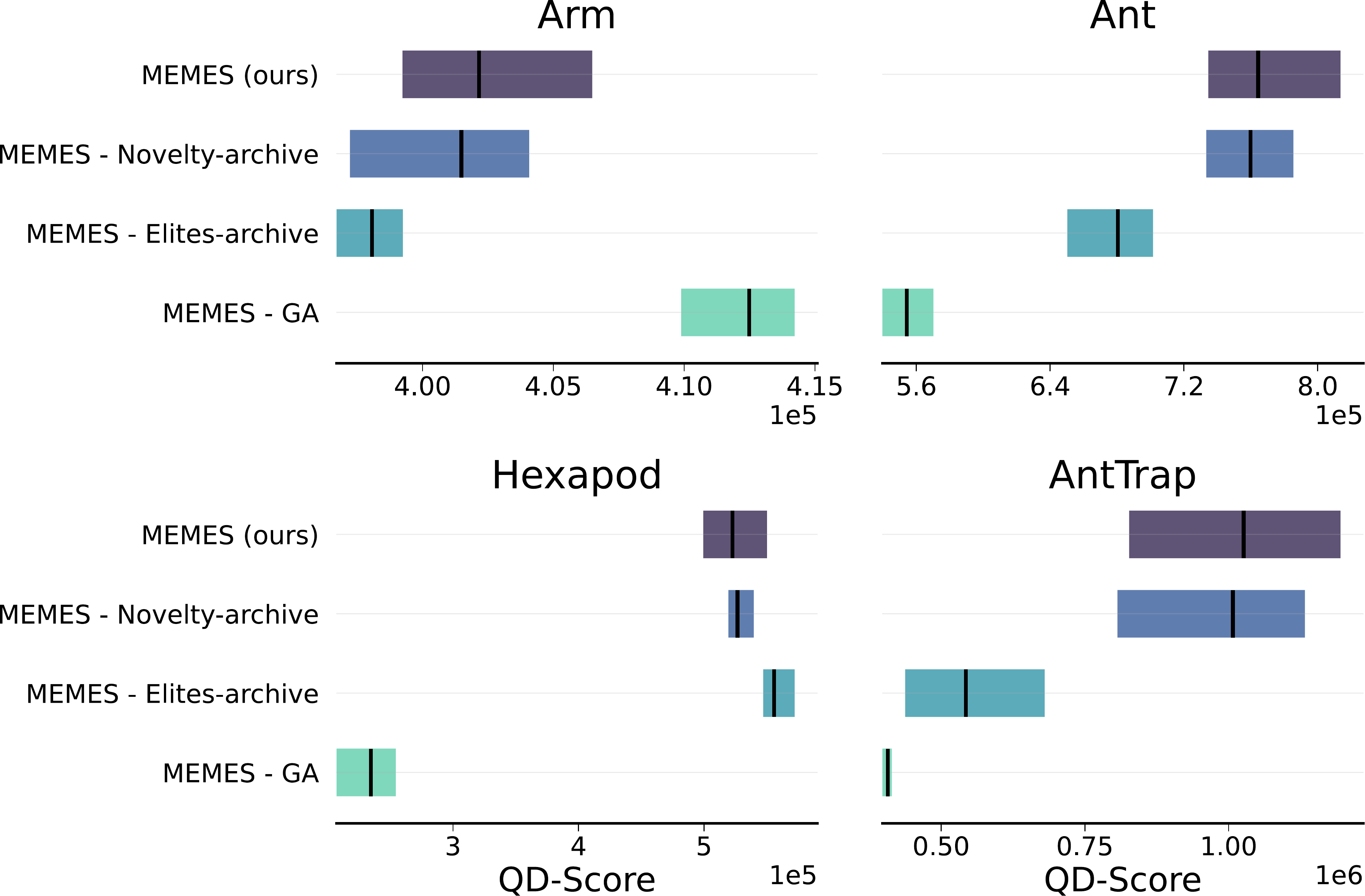}
\caption{
    Final QD-score for different novelty-computation mechanisms. We report the median and CI over $5$ seeds. 
}
\label{fig:novelty_result}
\end{figure}

Lastly, we study the different options for effective exploration (see Sec.~\ref{subsec:methods_explore_emitter}).
For fair comparison, we maintain the same ES-exploit emitter with the adaptive resets to isolate the performance of the different explore-emitter choices.
Figure~\ref{fig:novelty_result} shows the performance of the different explore emitters considered.
The fifo novelty archive performs similarly to the full novelty archive whilst being more memory efficient.
Interestingly, the archive of elites $\archive{}$ also represents another compute and memory efficient alternative that performs strongly on some tasks (i.e. Hexapod).
The decrease in performance in some tasks (especially deceptive) can be attributed to the loss of historical information by only maintaining elites as highlighted in Section~\ref{subsec:methods_explore_emitter}.
However, in tasks such as the Hexapod, where this information is less important and relevant, it performs the best.
The GA variant performs well on Arm but it clearly under-performs on more complex tasks, highlighting the benefit of the novelty ES emitters.

\section{Discussion and Conclusion}

In summary, we present \name{}, a QD algorithm that leverages tensorisation and large-scale parallelisation, to maintain multiple independent ES emitters (up to $\sim100$) in parallel, each with their own objective on a single GPU. 
\name{} also introduces an adaptive emitter reset mechanism to manage the parallel optimisation threads and maximise the improvement efficiency of the QD archive.
We demonstrate that \name{} outperforms a large variety of existing QD algorithms on a range of high-dimensional neuroevolution tasks and also lower-dimensional optimisation tasks.
Importantly, in addition to finding higher-performing and more diverse solutions, we also show that archive solutions generated by \name{} are more reproducible when operating in uncertain domains, making \name{} a more reliable QD algorithm.
This highlights the importance of effectively using samples and carefully considering them for addition in uncertain domains.

While \name{} performs consistently well, it is important to remember that this approach requires a large number of evaluations for estimating gradients. 
This limitation means that our proposed approach is currently effective only in settings where evaluations are cheap and can be parallelised on accelerator devices.
However, with powerful compute becoming cheaper, smaller, more widely available and heavy development and movement of environments on accelerated hardware~\cite{bonnet2023jumanji,jiang2023minimax, flair2023jaxmarl}, we believe \name{} can be an effective approach across a wide range of applications.
Given the flexibility of our algorithm, we hope more maximise the use of many parallel ES processes by extending \name{} to more objectives beyond fitness and novelty objectives in future work.

\newpage


\bibliographystyle{ACM-Reference-Format}
\bibliography{references}

\newpage
\newpage
\appendix

\section*{Appendix}

\section{Additional Details} \label{app:exp_details}

\subsection{Pseudo-code for ES and ME-ES} 

Algorithm~\ref{alg:es_optim} shows the optimisation procedure of the ES from \citet{salimans2017evolution} (see Sec.~\ref{subsec:ES}).
Algorithm~\ref{alg:mees} shows the procedure of the ME-ES algorithm by \citet{colas2020scaling} (see Sec.~\ref{subsec:ME-ES}).

\vspace{-2mm}
\SetCommentSty{mycommfont}
\SetKwProg{Def}{def}{:}{}
\begin{algorithm}[h]
\footnotesize
\DontPrintSemicolon
\caption{ES Optimization Step (Adapted from~\cite{salimans2017evolution})} 
\label{alg:es_optim}
    \textbf{Input:} objective $F$, sample size $N$, solution $\theta_t$
    
    \Def{\text{ES\_step}($F$, $N$, $\theta_t$)}{
    
        $\epsilon_1, ..., \epsilon_N \sim \mathcal{N}(0, I)$  \;

        \For{$i = 1, ..., N$}{
            $F_i = F(\theta_t + \sigma\epsilon_i)$ \;
        }
        
        $\theta_{t+1} \gets \theta_t + \alpha \frac{1}{N\sigma}\sum_{i=1}^{n}{F_i\epsilon_i}$ \;

        \Return $\theta_{t+1}$
    }
\end{algorithm}
\setlength{\textfloatsep}{2pt}

\vspace{-4mm}
\SetCommentSty{mycommfont}
\begin{algorithm}[h]
\footnotesize
\DontPrintSemicolon
\caption{MAP-Elites-ES (Adapted from~\cite{colas2020scaling})} 

\label{alg:mees}
    \textbf{Inputs:} sample size $N$, number of generations $N_{gen}$, $\optim{}$
    
    \textbf{Initialisation:} archive of elites $\archive{}$ and novelty-archive $\noveltyarchive{}$
    
    \For{$n_{gen} = 1, ..., N_{gen}$}{

        \uIf{$n_{gen} \mod (2 * \optim{}) == 0$}{
            \tcp{Explore mode}
        
            $\theta \gets \text{biased\_explore\_selection}(\archive{})$ \;

            $F \gets \text{novelty\_objective}(\noveltyarchive{})$ \;
        }
        \uElseIf{$(n_{gen} - \optim{}) \mod (2 * \optim{}) == 0$}{
            \tcp{Exploit mode}
        
            $\theta \gets \text{biased\_exploit\_selection}(\archive{})$ \;

            $F \gets \text{fitness\_objective}()$ \;
        }
        
        \tcp{Apply ES}
        $\tilde\theta \gets\text{ES\_step}(F, N, \theta) $ \;
        

        \tcp{Evaluate and update archives}
        $\archive{}, \noveltyarchive{} \gets \text{update\_archives}(\tilde\theta)$ \;

    }
\Return $\archive{}$
\end{algorithm}
\setlength{\textfloatsep}{2pt}

\subsection{Uniform Selection during Emitter Reset} 

Another key design of ME-ES is its biased sampling strategy. Each time the ME-ES emitter is re-initialized, a new parent is sampled from the archive using a complex biased selection strategy.
It involves selecting cells with a high fitness for the exploit mode and cells with a high novelty-score for the explore mode.
Thus, it assumes that high-fitness parents may lead to higher-fitness offspring and similarly for high-novelty.
Based on this assumption, this selection strategy seems a reasonable choice given that only one parent is selected at every $\optim{}$ generation and that solutions are only optimised for $\optim{}$ steps before reset.

However, while this assumption may generally be true, it is not always the case. One example is deceptive-rewards scenarios~\cite{lehman2011evolving}. 
Thus, this selection procedure could end up negatively impacting the efficiency of the algorithm.
Additionally, a common hypothesis in QD is that uniform selection promotes the use of stepping stones in the archive to obtain even more promising solutions \cite{mouret2015illuminating, gaier2019quality}.
Thus, the original ME algorithm used uniform selection. 
Hence, for \name{} we also choose uniform selection to reset the ES emitters.

\subsection{Tasks and Environments}

To evaluation our method, we purposefully select four distinctly different tasks:
Arm, Hexapod, Ant and AntTrap.
The Ant and AntTrap tasks are RL-based neuroevolution tasks in which the search space corresponds to the parameters of a neural network policy.
On the other hand, the Hexapod and Arm represent more mathematical optimisation tasks. We provide details of each environment below:

\subsubsection{Arm}
This task involves a n-DoF planar robotic arm where the objective is to find the all the reachable $(x, y)$ positions in the end-effector workspace of the robot, that are as smooth as possible. This is also known as the inverse-kinematic problem in robotics. 
The search space in this task corresponds to the relative joint angles for each DoF of the arm.
As the arm is redundant, there can be multiple solutions for an $(x, y)$ end-effector position.
To find smooth solutions, the objective $F(\theta)$ is defined to minimize the variance of the joint positions.
The feature space then corresponds to the $(x,y)$ positions of the end-effector.
The arm task is used as a simple and fast benchmark task.
We select this task to also demonstrate that our method can be applied to general QD optimisation problems that do not have to be an MDP/RL or neuroevolution problem. 
The search space, feature space and fitness values all lie in the range $[0,1]$.
Equations for the $d$ and $F$ are given below where $FK$ is the 2-D Forward Kinematics equation:

\vspace{-2mm}
\begin{align}
    d(\theta) &=
    \begin{pmatrix}
    FK_x(\theta)\\
    FK_y(\theta)
    \end{pmatrix}\\
    F(\theta) &= 1 - \sqrt{
    \frac{1}{n}\sum_{i=1}^{n}{(\theta_i - \tilde{\theta})^2}
    }
\end{align}

\subsubsection{Hexapod}
The hexapod robot is an 18-DoF robot with 12 independent DoF. The joint angle trajectory follows a cyclical sinusoidal pattern and is parameterised by two values, the amplitude and phase of a sinusoidal wave~\cite{cully2015robots}.
The fitness $F(\theta)$ is defined by the angular yaw of the robot following a circular trajectory while the feature $d(\theta)$ corresponds to the final $(x, y)$ position of the robot after $T$ seconds~\cite{cully2013behavioral}.
This task is used to evaluate \name{} on a more complex non-MDP/RL tasks.
Equations for the $d$ and $F$ are given below, where $\alpha(\theta)$ is the final orientation of the robot and $\beta(\theta)$ the tangent of the circle through its start and end position: 

\vspace{-2mm}
\begin{align}
    d(\theta) &= 
    \begin{pmatrix}
    x_T \\
    y_T \\
    \end{pmatrix}\\
    F(\theta) &= -| \alpha(\theta) - \beta(\theta) |
\end{align}

\begin{table*}[t!]
\footnotesize
\centering
\begin{tabular}{l|ccccccccc}
\toprule
\textsc{Hyperparameter} & \textsc{ME} & \textsc{PGA-ME} & \textsc{ES} & \textsc{NS-ES} & \textsc{NSR-ES} & \textsc{NSRA-ES} & \textsc{ME-ES} & \textsc{CMA-ME} & \name{} \\ 

\midrule
\addlinespace

Iso coefficient, $\sigma_1$  & 0.01 & 0.01 & & & & & & & \\
Line coefficient, $\sigma_2$ & 0.1 & 0.1 & & & & & & & \\

\midrule
\addlinespace

Samples sigma & & & 0.02 & 0.02 & 0.02 & 0.02 & 0.02 & 0.5 & 0.02 \\
Num. of samples & & & 512 & 512 & 512 & 512 & 10000 & & 512 \\
Learning rate & & & 0.01 & 0.01 & 0.01 & 0.01 & 0.01 & & 0.01 \\
Num. nearest neighbors & & & & 10 & 10 & 10 & 10 & & 10 \\
l2 normalisation coefficient & & & & 0 & 0 & 0 & 0.01 & & 0 \\

\midrule
\addlinespace

Reset budget $S_{max}$ & & & & & & & & & 32 \\
Proportion explore $p_{explore}$ & & & & & & & & & 0.5 \\
Fifo size & & & & & & & & & 50000 \\

\midrule
\addlinespace

Pop. size & & & & 5 & 5 & 5 & & & \\
Fitness weight & & & & & 0.5 & 1.0 & & & \\
Adapt. fitness weight amount & & & & & & 0.05 & & & \\
Adapt. fitness weight period & & & & & & 50 & & & \\

\midrule
\addlinespace

Pool size & & & & & & & & 15 & \\
Emitter type & & & & & & & & Imp. & \\

\midrule
\addlinespace

Proportion PG $p_{PG}$ & & 0.5 & & & & & & & \\ 
Num. PG training steps & &  100 & & & & & & &  \\ 
Replay buffer size & & $10^6$ & & & & & & & \\
Policy hidden layer sizes & & [64, 64] & & & & & & & \\
Critic hidden layer size & &  [256, 256] & & & & & & & \\
Critic learning rate & & 0.0003 & & & & & & &  \\ 
Greedy learning rate &  & 0.0003 & & & & & & & \\ 
PG variation learning rate & &  0.001 & & & & & & & \\
Transitions batch size & &  256 & & & & & & & \\
Noise clip & & 0.5 & & & & & & & \\
Policy noise & & 0.2 & & & & & & & \\
Soft $\tau$ update & &0.005 & & & & & & & \\

\bottomrule
\end{tabular}
\caption{
    Hyperparameters of baselines considered in our experiments.
}
\label{tab:hyperparams}
\end{table*}

\subsubsection{Ant} The Ant task is a QD benchmark task~\cite{cully2015robots, flageat2022benchmarking} where the objective is to find a set of diverse gaits for moving forward as fast as possible.
The solutions found in such an archive are shown to be effective for damage adaptation~\cite{cully2015robots}.
The fitness $F(\theta)$ is defined by the forward velocity of the Ant while minimising the torque used (to represent efficient walking gaits).
The feature $d(\theta)$ is defined by the average time each foot is in contact with the ground, a diversity which results in diverse walking gaits.
Here, we consider the uncertain case where the initial position and velocity of the joints are randomly sampled at the start of each episode, making the task a good example of a UQD task (see Sec.~\ref{subsec:uqd}).
Equations for the $d$ and $F$ are given below:

\vspace{-2mm}
\begin{align}
    d(\theta) &= \frac{1}{T} \sum_{t}^{T}{
    \begin{pmatrix}
    C_1^\theta(t) \\
    \vdots \\
    C_I^\theta(t) 
    \end{pmatrix}\textrm{, where $I$ is the number of feet.}
    } \\
    F(\theta) &= \sum_{t=0}^{T}{r_{forward}(\theta) + r_{survive}(\theta) + (-r_{torque}(\theta))}
\end{align}

\subsubsection{AntTrap}
The AntTrap task is categorised as a hard exploration QD benchmark task~\cite{chalumeau2022assessing} as it consists of deceptive rewards. 
The goal of this task is to move forward as fast as possible with the caveat of there being a trap in front of the ant.
The fitness $F(\theta)$ is defined similarly to the Ant as a combination of the forward velocity and energy used while the feature $d(\theta)$ is defined by the final $(x, y)$ position of the robot at the end of the episode.
As mentioned in Section~\ref{subsec:exp-setup}, the nature of this task differs significantly from the Ant described above, despite sharing the same Ant robot morphology for simplicity.
Similarly to the Ant task, we consider the uncertain case with random initial positions. To the best of our knowledge, no other work study this complex UQD task that requires solutions to go around the trap in a reproducible manner, making it highly challenging.
Equations for the $d$ and $F$ are given below:

\vspace{-2mm}
\begin{align}
    d(\theta) &= 
    \begin{pmatrix}
    x_T \\
    y_T \\
    \end{pmatrix}\\
    F(\theta) &= \sum_{t=0}^{T}{r_{forward}(\theta) + r_{survive}(\theta) + (-r_{torque}(\theta))}
\end{align}

\begin{figure*}[h]
    \centering
    \includegraphics[width=\hsize]{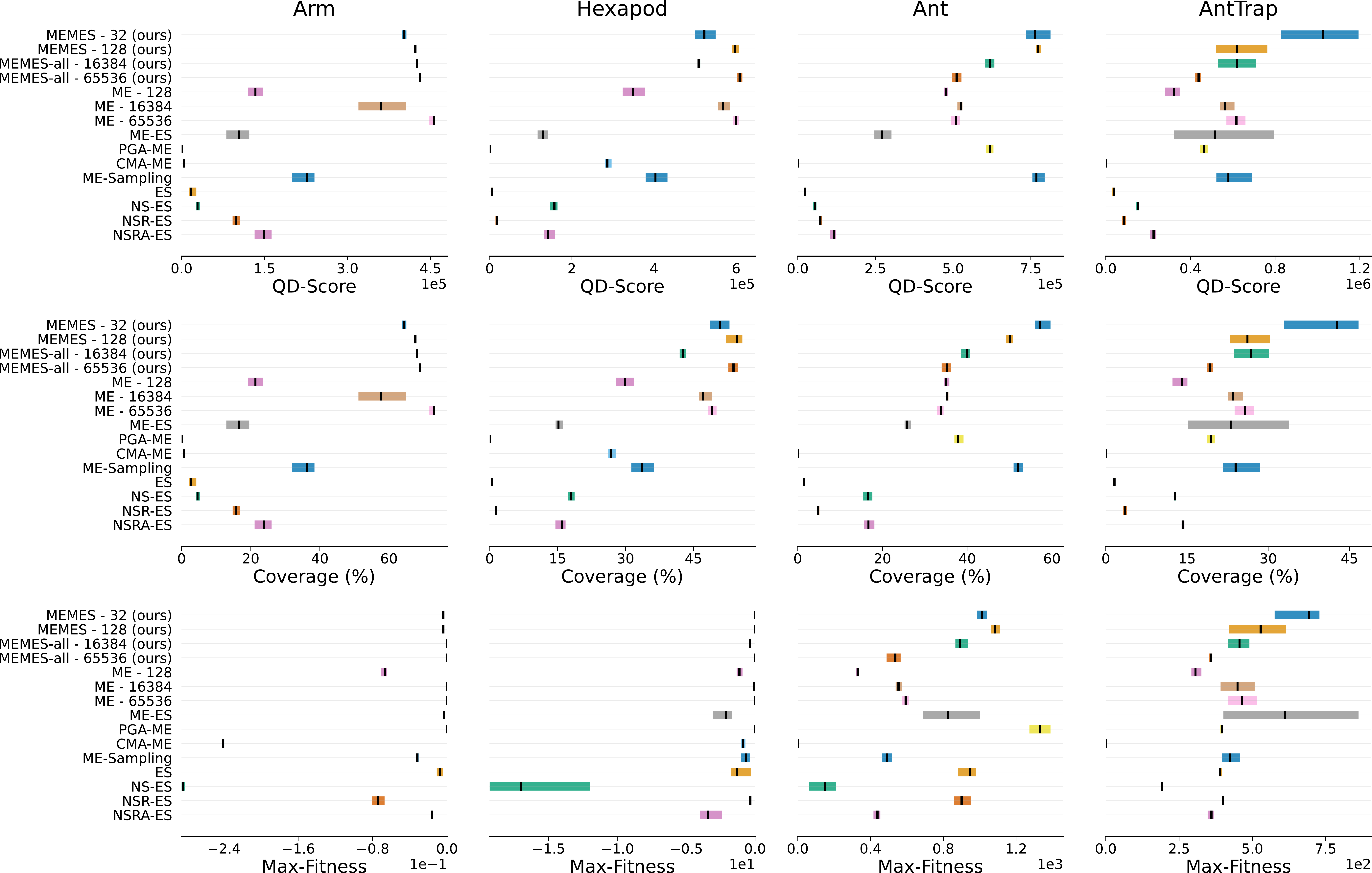}
    \caption{
    Final QD-Score (top), Coverage (middle) and Max-Fitness (bottom). We report the median and CI over $10$ seeds.
    }
    \label{fig:appendix}
\end{figure*}

\section{Implementation and hyperparameters}

The implementation of all our baselines is part of the QDax library. We give in Table~\ref{tab:hyperparams} the hyperparameters used for each baseline.

\section{Supplementary results} \label{app:more_results}

\subsection{MEMES with larger batch-size} 

To demonstrate the ability of MEMES to scale to higher numbers of parallel ES, we run MEMES and MEMES-all with $128$ parallel ES. 
For a fair comparison, we also re-run ME with batch size $65536$, which corresponds to the number of evaluations used by MEMES - $128$. 
All of these variants still run on a single commonly available GPU device. However, due to time and computational constraints, these variants were only run on the AntTrap for $4000$ generations instead of $10000$ for the main approaches. The addition of $65536$ solution to the ME archive, as well as the computation of novelty for $65536$ individuals both required to be batched to fit on the RAM of our GPU, significantly increasing the run time and leading to this limitation.
Hence, we find MEMES - $32$ to be a good parameter that balances runtime and good performance on these tasks but scaling to MEMES - $128$ and larger could be important and critical in more complex and open-ended task settings that necessitate a large number of independent optimisation threads.

We display the new comparison in Figure~\ref{fig:appendix}. The results show that MEMES - $128$ is also competitive, managing to get even better results on the Arm, Hexapod and Ant tasks. 
Its significantly lower performance on the AntTrap can be fully attributed to the lower number of generations it has been run for. 

\subsection{Time comparison} 

Table~\ref{tab:time} shows the run-time comparison of the algorithms. This shows that despite using significantly more evaluations and performing significantly better, MEMES only requires marginally more run-time than our baselines using the same number of evaluations.
All experiments were done on a single GPU as well for fair comparison also demonstrating that our implementation of MEMES can handle $32$ parallel ES processes at an efficient run-time.

This marginal increase in runtime can be attributed mainly to operations related to computing empirical gradients.
In particular, the novelty score computation which while has already been significantly sped up using the fifo mechanism proposed, still requires the computation of thousands of distances between the ES samples and the novelty archive content.
This is confirmed by a $\sim 9\%$ increase in runtime compared to ME with a batch size of $16384$ (same number of total evaluations as \name{} per generation). 
Additionally, despite the massive parallelisation of evaluations, these evaluations still contribute to additional runtime, as highlighted when comparing ME-$128$ with ME-$16384$.

\begin{table}[h]
\footnotesize
\centering
\begin{tabular}{l|ccc}
\toprule
\textsc{Time (mins)}
 & \textsc{Hexapod}
 & \textsc{Ant}
 & \textsc{AntTrap}
\\ 

\midrule
\addlinespace

\textsc{ME}
& 4
& 25
& 81
\\

\textsc{ME - 16384}
& 60
& 69
& 753
\\

\textsc{PGA-ME}
& -
& 150
& 383
\\

\textsc{MEMES}
& 64
& 77
& 813
\\

\textsc{MEMES-all}
& 87
& 100
& 960
\\

\bottomrule
\end{tabular}
\caption{
        Average wall-clock time comparison on the same hardware, in minutes.
        We purposefully do not report the time for Arm since this task is extremely fast to run.
}
\label{tab:time}
\end{table}.

\subsection{Convergence Results}

\begin{figure*}[t!]
\centering
\includegraphics[width = \hsize]{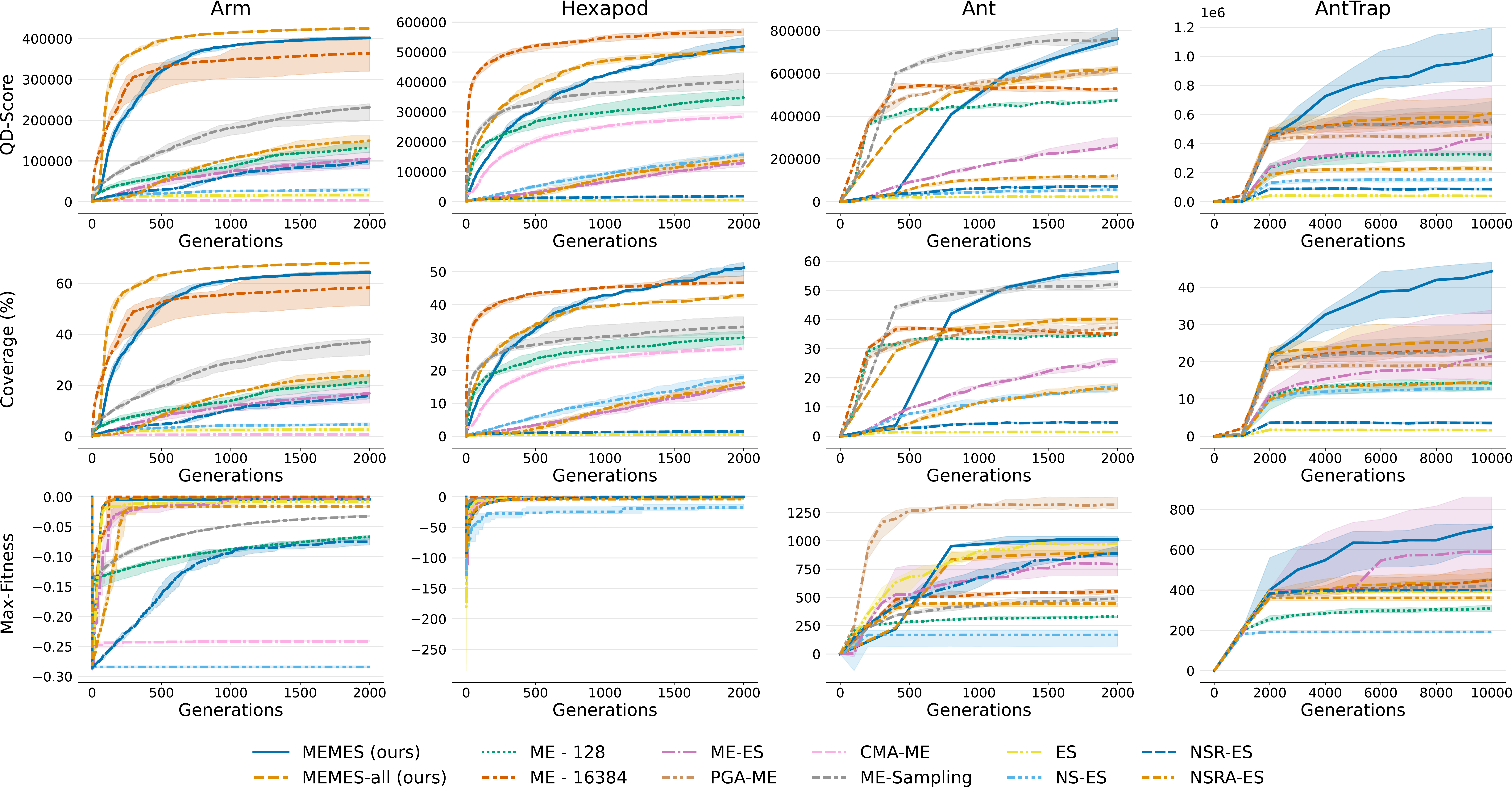}
\caption{
    QD-score (top), Coverage (middle) and Max-fitness (bottom) across generations.
    The solid line is the median and the shaded areas are the quartiles over $5$ replications.
}
\label{fig:convergence}
\end{figure*}

\begin{figure*}[t!]
\centering
\includegraphics[width = \hsize]{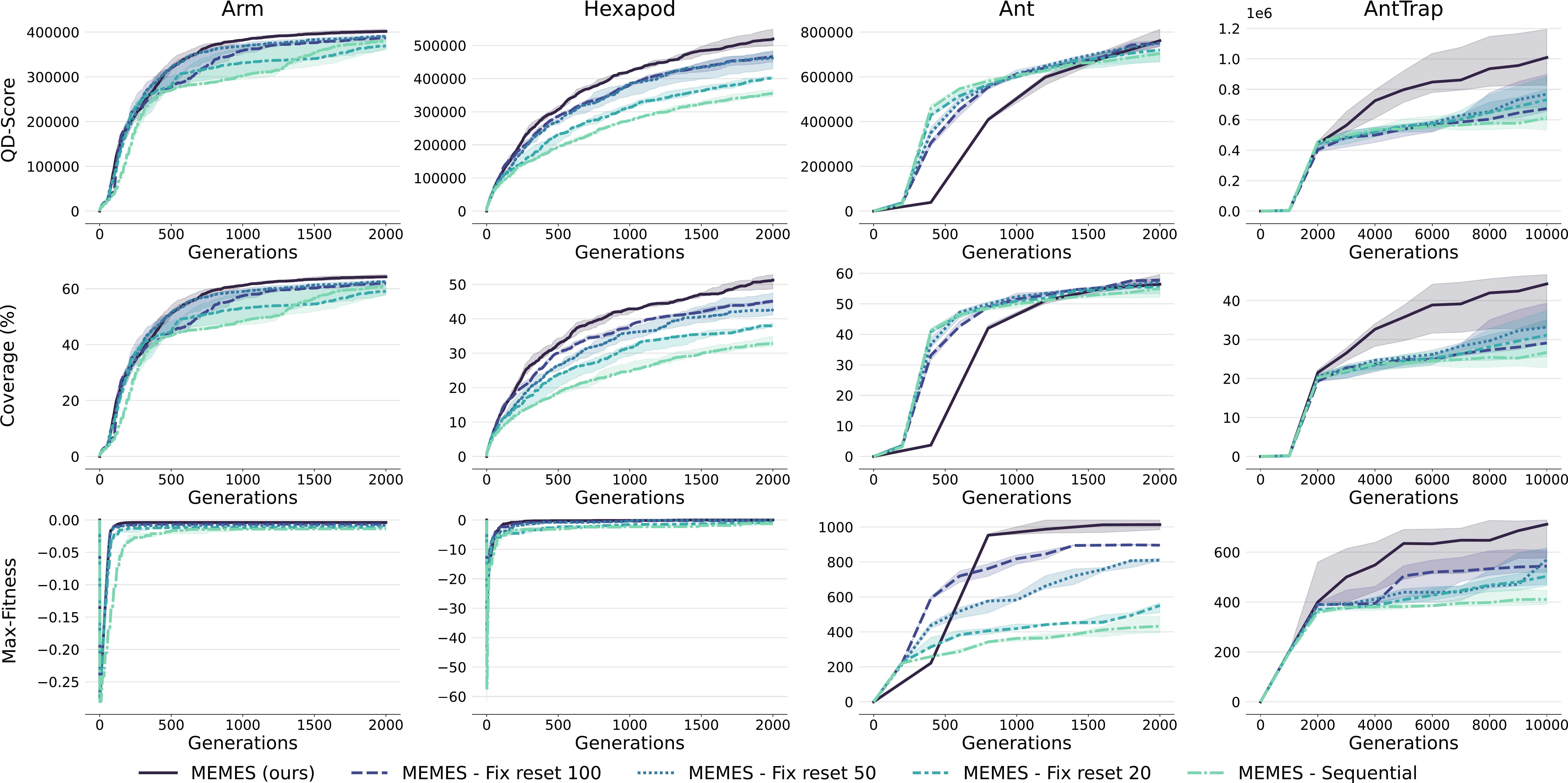}
\caption{
    QD-score (top), Coverage (middle) and Max-fitness (bottom) for different fixed reset values, compared to the adaptive reset mechanism across generations.
    The solid line is the median and the shaded areas are the quartiles over $5$ replications.
}
\label{fig:reset_ablation_convergence}
\end{figure*}

\begin{figure*}[t!]
\centering
\includegraphics[width = \hsize]{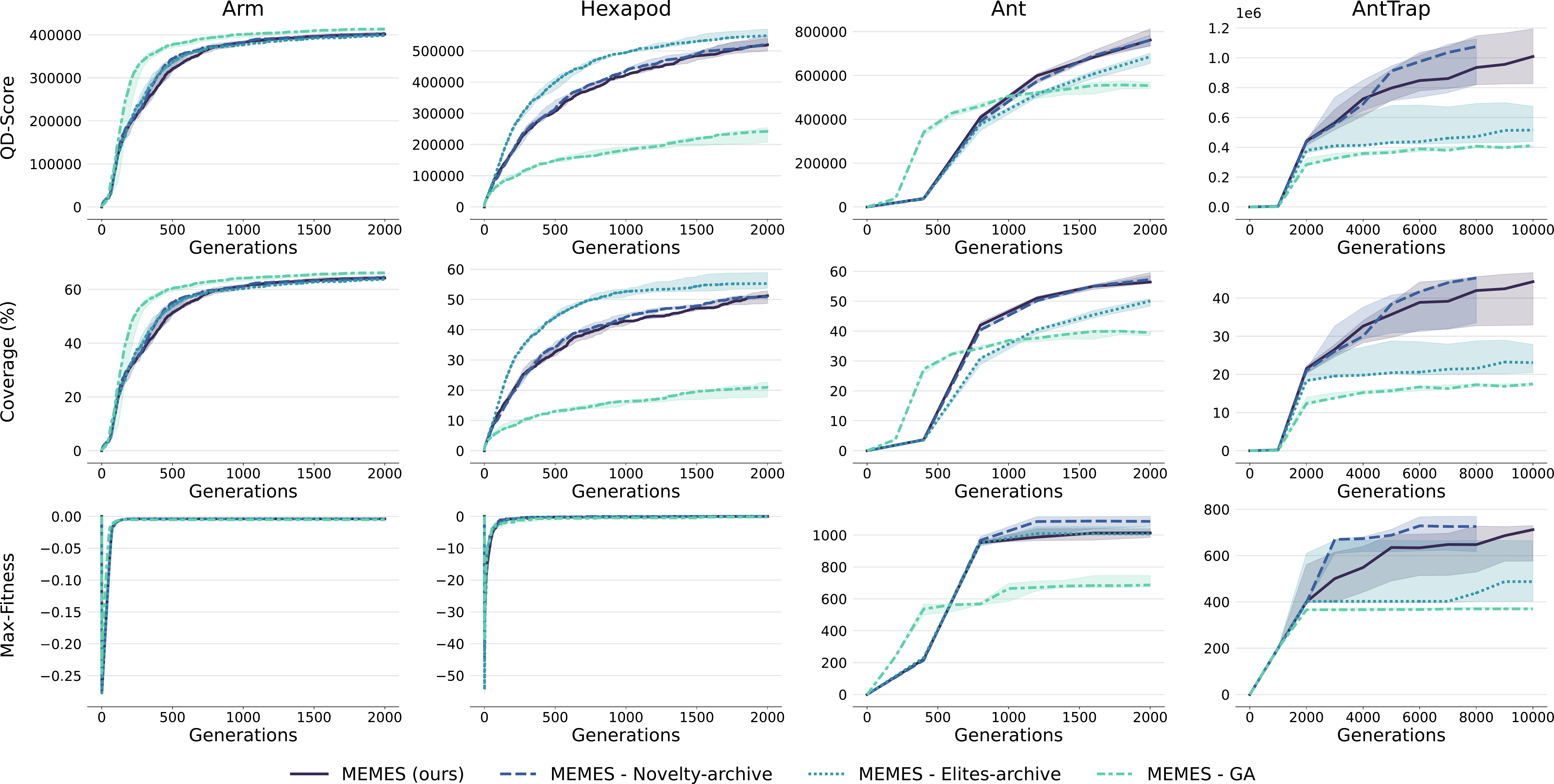}
\caption{
    QD-score (top), Coverage (middle) and Max-fitness (bottom) for different novelty-computation mechanisms across generations.
    The solid line is the median and the shaded areas are the quartiles over $5$ replications.
}
\label{fig:novelty_ablation_convergence}
\end{figure*}

We provide in Figure~\ref{fig:convergence},~\ref{fig:reset_ablation_convergence} and~\ref{fig:novelty_ablation_convergence} the convergence plots (performance metrics across generations) for the main results, the \name{} reset ablation and the \name{} novelty ablation respectively.

\subsection{Archive plots} 

\begin{figure*}
    \centering
    \includegraphics[width=0.7\hsize]{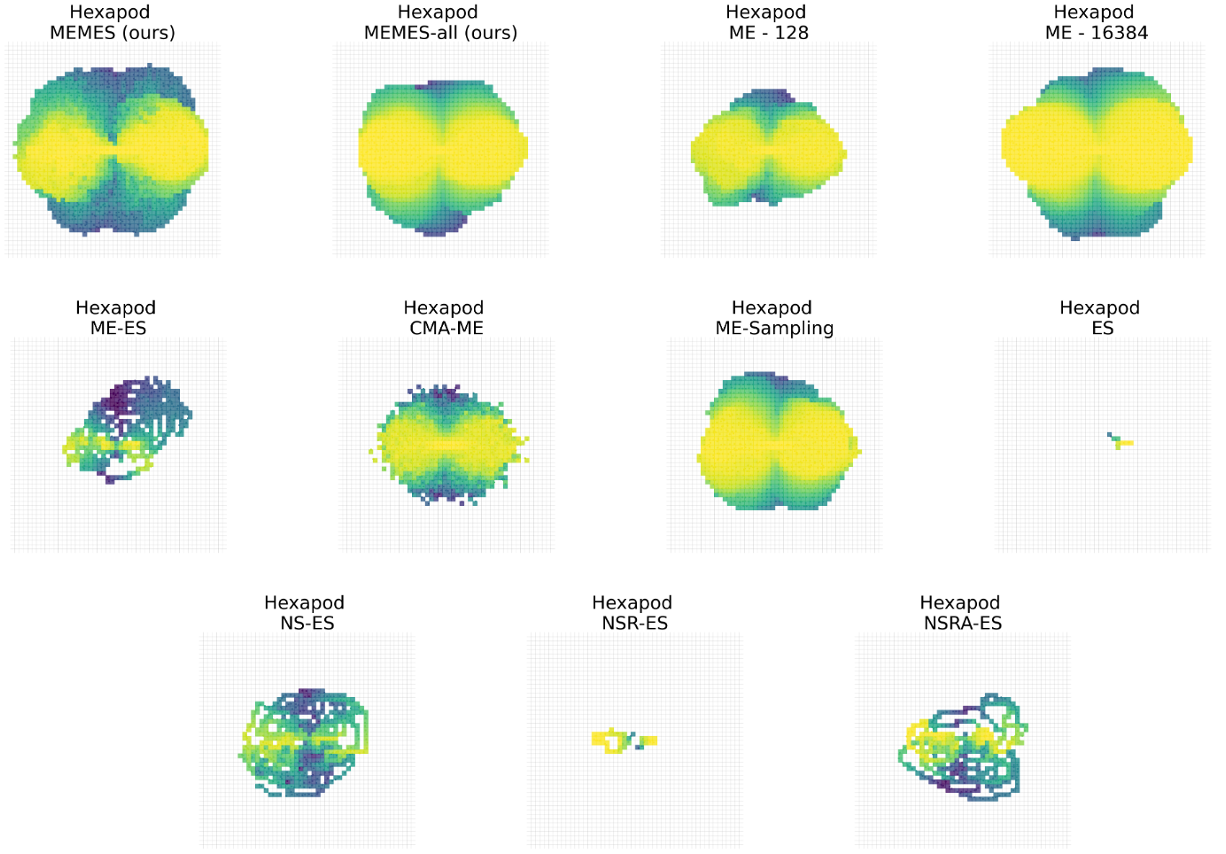}
    \caption{
    Hexapod archives.
    }
    \label{fig:hexapod}
\end{figure*}

\begin{figure*}
    \centering
    \includegraphics[width=0.7\hsize]{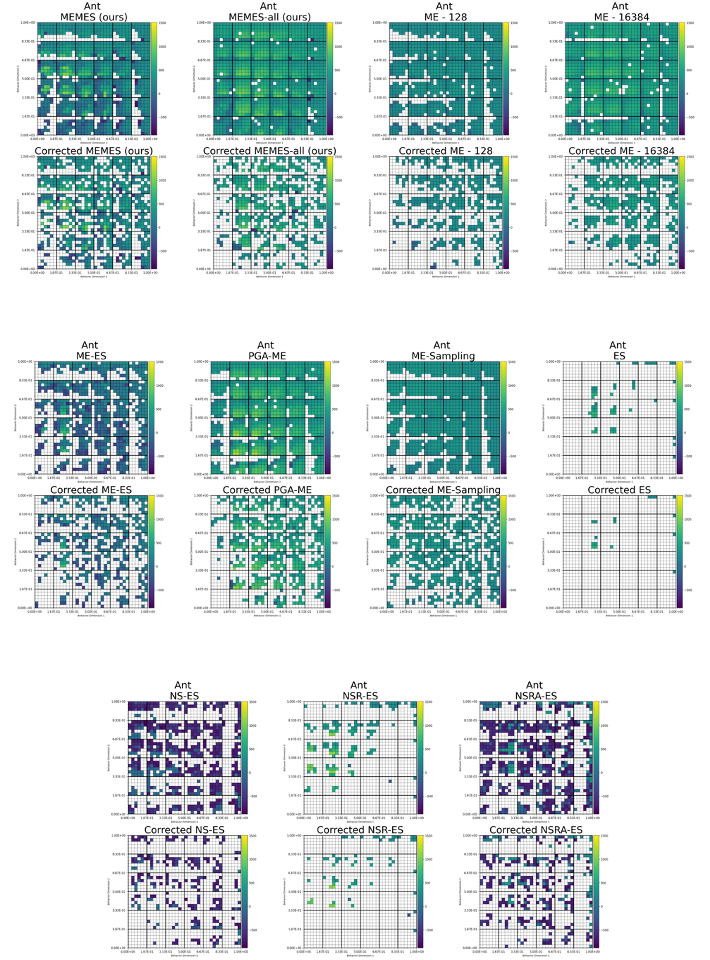}
    \caption{
    Ant archives: we display both the illusory archives (top) and the corrected archives (bottom).
    }
    \label{fig:ant}
\end{figure*}

\begin{figure*}
    \centering
    \includegraphics[width=0.7\hsize]{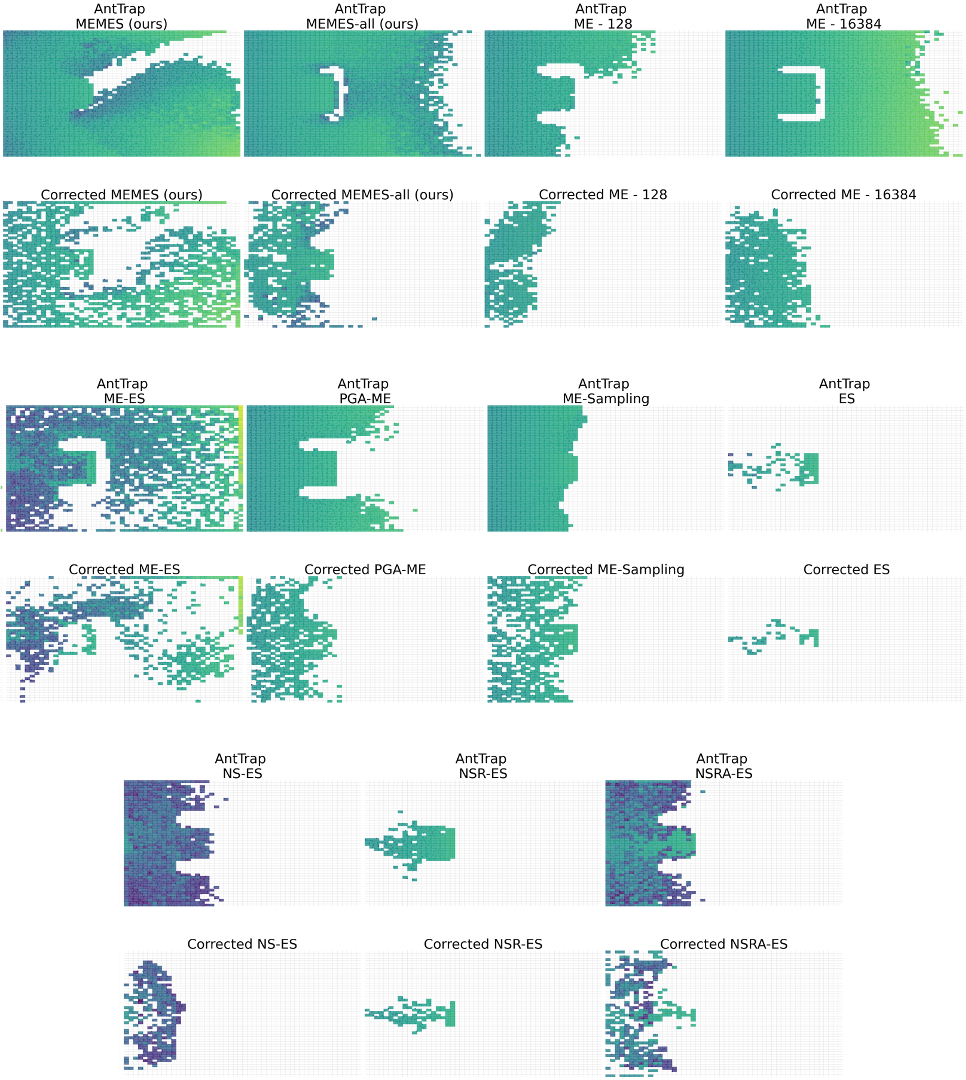}
    \caption{
    AntTrap archives: we display both the illusory archives (top) and the corrected archives (bottom).
    }
    \label{fig:anttrap}
\end{figure*}

Figure~\ref{fig:anttrap} and ~\ref{fig:ant} display the archives for the uncertain AntTrap and Ant uncertain neuroevolution tasks respectively.
Figure~\ref{fig:hexapod} displays the archive plots for the Hexapod optimisation task.

\end{document}